\documentclass[11pt]{article}

\usepackage[final]{acl}

\usepackage{times}
\usepackage{latexsym}
\usepackage[T1]{fontenc}
\usepackage[utf8]{inputenc}
\usepackage{microtype}
\usepackage{inconsolata}
\usepackage{graphicx}
\usepackage{booktabs}
\usepackage{amsmath}
\usepackage{tcolorbox}
\usepackage{longtable}
\usepackage{xcolor}

\title{Load-Bearing Context: The Question Damage Score for Evaluating Context Reliance in Linguistic Reasoning}

\author{Neh Majmudar \\
  CUNY \\
  \texttt{neh.majmudar41@gc.cuny.edu} \\\And
  Elena Filatova \\
  CUNY \\
  \texttt{efilatova@citytech.cuny.edu} \\}

\begin{document}
\maketitle

\begin{abstract}

Determining whether large language models (LLMs) derive their answers from the provided context or from prior knowledge remains a fundamental challenge. Linguistic Rosetta Stone puzzles from linguistic olympiads provide a uniquely controlled setting: each puzzle is self-contained, and all answers can be derived solely from a small set of expert-designed context examples without requiring external knowledge. Removing individual context examples, however, may eliminate information required to answer one or more questions while leaving the remainder of the puzzle unchanged. We leverage this property to introduce a diagnostic framework for analyzing the role of individual context examples in puzzle solving. Starting from 53 curated puzzles from the United Kingdom Linguistics Olympiad (UKLO), we generate two modified variants of each puzzle by deleting a single context example: (1) uniform random deletion and (2) targeted deletion inspired by error-correcting code (ECC) theory designed to remove a structurally load-bearing example uniquely carrying information required for answering at least one question. We formalize the impact of these deletions with a \textbf{Question Damage Score} ($D_Q$), which classifies puzzles as \emph{fragile} or \emph{robust} prior to model evaluation. As a first application of the framework, we evaluate three frontier LLMs on the original and modified puzzles under explicit instructions to abstain when the provided information is insufficient. Despite these instructions, the models rarely abstain and often continue producing correct answers after structurally load-bearing context is removed. These findings motivate further investigation into the roles of context-based reasoning, prior knowledge, memorization, and linguistic inference. Beyond abstention, the proposed framework enables fine-grained analyses of context reliance, including causal interventions, stopping-set analysis, targeted contamination studies, and mechanistic interpretability.

\end{abstract}

\section{Introduction}
\label{sec:intro}

Large language models (LLMs) are typically evaluated on benchmarks in which every problem has a correct answer. While such benchmarks measure task accuracy, they reveal little about \textbf{how} models arrive at their answers or whether those answers are derived from the provided context rather than prior knowledge or memorization. This limitation has motivated recent work on unanswerable benchmarks, where models are expected to recognize insufficient information and abstain from answering~\citep{AbstentionBench,madhusudhan-etal-2025-llms}. Most existing approaches, however, generate unanswerable instances by removing the entire problem context or by constructing inherently unanswerable questions, making it difficult to analyze the contribution of individual information pieces.

Rosetta Stone puzzles from the United Kingdom Linguistics Olympiad (UKLO) provide a uniquely controlled setting for studying context reliance. Each puzzle consists of a small set of examples in one language paired with their English translations, followed by questions that require the solver to infer the underlying linguistic rules solely from the provided examples and apply those rules to previously unseen utterances. The puzzles are self-contained: all information needed to solve them is within the provided examples, no external linguistic knowledge is needed. Examples of UKLO Rosetta Stone puzzles are provided in App.~\ref{sec:appExamples}.

We introduce a diagnostic framework for studying the role of individual context examples in puzzle solvability and the effect of removing those examples on LLM behavior. Starting from 53 UKLO puzzles, we generate two modified variants of each puzzle by deleting a single context example: (1) uniform random deletion and (2) targeted deletion inspired by error-correcting code (ECC) theory, which targets a load-bearing context example that uniquely carries information required to answer at least one question. We formalize the impact of these deletions using the \textbf{Question Damage Score} ($D_Q$), which counts the number of questions that become unanswerable after removing context example $l_j$. A puzzle with $\max_j D_Q(j)>0$ contains at least one load-bearing context example whose removal renders one or more questions structurally unanswerable; we call such puzzles \emph{fragile}. In contrast, a puzzle with $\max_j D_Q(j)=0$ contains sufficient redundancy that removing any single context example preserves the solvability of every question; we call such puzzles \emph{robust}.

As a first application of the framework, we investigate LLM abstention. Three frontier LLMs solve the original and modified puzzles under explicit instructions to abstain whenever the provided context is insufficient. Across all models, abstention is rare, even after structurally important context is removed, and models often answer correctly on structurally unsolvable variants. Rather than explaining these answers, our framework identifies cases where reasoning over the remaining context alone appears insufficient, motivating future studies of context-based reasoning, prior knowledge, memorization, and linguistic inference.

Although we focus on abstention, the framework is more general. By identifying structurally important context examples and enabling controlled perturbations, it provides a foundation for causal interventions, interpretability, targeted contamination studies, and broader analyses of context reliance in reasoning tasks.

Our contributions are:

\begin{enumerate}

\item \textbf{A diagnostic framework} for controlled, single-example context perturbation of self-contained reasoning tasks, including uniform random and ECC-inspired targeted deletion together with the \textbf{Question Damage Score} ($D_Q$), which identifies structurally load-bearing examples and classifies puzzles before model evaluation. 

\item \textbf{An empirical case study} demonstrating that frontier LLMs rarely abstain after structurally important information has been removed, revealing a disconnect between structural puzzle solvability and model behavior. 

\item \textbf{A reusable evaluation methodology} that extends beyond abstention and supports future studies of context reliance, memorization, causal analysis, mechanistic interpretability. 

\end{enumerate}


\section{Related Work and Positioning}
\label{sec:related}

\subsection{Unanswerability and Abstention in LLMs}

Recent work has investigated whether LLMs can recognize when the provided information is insufficient to answer a question. Existing benchmarks typically evaluate this ability by constructing inherently unanswerable questions, removing the supporting context, or introducing missing information, expecting models to abstain rather than hallucinate answers. For example, Abstain-QA introduces multiple-choice questions with no correct answer option~\citep{madhusudhan-etal-2025-llms}. SQuAD 2.0 augments the original SQuAD dataset with adversarially constructed unanswerable questions~\citep{SQuAD}. The SelfAware benchmark~\citep{yin2023largelanguagemodelsknow} targets questions that are inherently unanswerable: speculative, subjective, or lacking scientific consensus. These studies consistently show that modern LLMs frequently produce confident answers despite lacking sufficient evidence.

In these benchmarks, unanswerable items are constructed independently of any answerable counterpart. Our modified puzzles are derived from their originals by a single deletion, so each unanswerable variant has a matched solvable counterpart differing in exactly one context example.

\subsection{Perturbation for Reasoning Analysis}

Several approaches generate unanswerable problems by removing information. For example, GSM8K-Abstain~\citep{AbstentionBench} removes the entire problem context, making each resulting question unanswerable (see App.~\ref{sec:appTreeCut}, Fig.~\ref{fig:ContextRemoval}).

TreeCut~\citep{ouyang-2025-treecut} studies context dependence in synthetically generated arithmetic problems, each generated from a dependency tree of variables and arithmetic relations. The structure is available by construction, thus, the solution follows a unique path of basic arithmetic operations, unanswerable variants can be produced by removing an edge along that path, and unsolvability is decidable (see App.~\ref{sec:appTreeCut}, Fig.~\ref{fig:TreeCut}). 

Linguistic puzzles lack an explicit derivation tree: dependencies between context examples and questions must be recovered from surface form. This makes our structural criterion approximate rather than exact, but it permits analysis of published problems, where recall from pretraining is a live possibility rather than one excluded by design.

\subsection{Linguistic Olympiad Puzzles as Self-Contained Reasoning Tasks}

Linguistic Olympiad puzzles have recently emerged as valuable benchmarks for evaluating LLM reasoning. Such puzzles require models to infer linguistic rules from examples rather than rely on prior knowledge~\citep{sahin-etal-2020-puzzling,linguinibenchmarklanguageagnosticlinguistic,LingOly,chi-etal-2024-modeling, cunveiling}. 
Perturbation-based variant, LingOly-TOO~\citep{khouja2026lingolytoodisentanglingreasoningknowledge}, alters surface form to control for memorization. 

Rosetta Stone puzzles are particularly well suited for controlled context perturbation~\citep{derzhanski2010linguistic,bozhanov-derzhanski-2013-rosetta}. Each puzzle is self-contained: all linguistic rules can be inferred from a small collection of expert-designed context examples, and no external linguistic knowledge is required. Consequently, individual context examples can be removed while preserving the remainder of the puzzle, enabling systematic analysis of how missing evidence affects both puzzle solvability and model behavior.

Our contribution is neither a new collection nor a surface transformation, but a set of controlled deletions that remove information and record which questions depend on it.

\section{Dataset Overview}
\label{sec:dataset}
\label{sec:dataset}

Rosetta Stone puzzles from the United Kingdom Linguistics Olympiad (UKLO) are widely used as a benchmark for evaluating compositional reasoning in LLMs as they require inferring linguistic rules from a small set of examples without relying on prior knowledge of the target language~\citep{linguinibenchmarklanguageagnosticlinguistic,khouja2026lingolytoodisentanglingreasoningknowledge}. Each puzzle contains a set of context examples, which provide evidence for discovering the underlying patterns, and a set of questions to be solved by applying the discovered linguistic patterns. Since all information required to solve the puzzle is contained within the provided examples, Rosetta Stone puzzles provide a self-contained reasoning environment. See App.~\ref{sec:appExamples} for example UKLO Rosetta Stone puzzles. 

In this work, we analyze 53 Rosetta Stone puzzles selected from 106 UKLO puzzles published between 2010 and 2025. We exclude those puzzles whose structure does not support controlled single-example perturbations, including puzzles with pictorial context (App.~\ref{sec:appExamples}, Fig.~\ref{fig:Chinese}), puzzles requiring multiple situational perspectives (App.~\ref{sec:appExamples}, Fig.~\ref{fig:Longgu}), and puzzles involving more than two languages (App.~\ref{sec:appExamples}, Fig.~\ref{fig:Korowai}). The resulting corpus contains 53 puzzles with an average of 16 context examples ($\min=6$, $\max=38$). See Sec.~\ref{sec:DataAndCode}.

The relatively small corpus is a deliberate trade-off between scale and curation quality. Each puzzle is a carefully curated reasoning task: a linguist designs the underlying grammatical phenomenon and supporting examples, the UKLO committee reviews the puzzle for correctness and difficulty, and participant performance provides an empirical measure of difficulty. Each puzzle contains multiple context examples and evaluation questions, enabling multiple controlled perturbations.

The modular organization of Rosetta Stone puzzles makes them well suited for controlled context perturbations. Individual context examples can be removed while preserving the reasoning task, enabling us to distinguish consequential deletions from inconsequential ones. Unlike previous work, which primarily uses these puzzles as reasoning benchmarks~\citep{sahin-etal-2020-puzzling,chi-etal-2024-modeling,LingOly}, we exploit their structure to characterize the contribution of individual context examples independently of model behavior.

\section{Deletion as a Diagnostic Probe}
\label{sec:deletion_strategies}
\label{sec:deletion}

Using the UKLO Rosetta Stone corpus (Section~\ref{sec:dataset}), we introduce a controlled context perturbation framework that removes a single context example while preserving the remainder of the reasoning task. Whether the modified puzzle remains solvable depends on the information carried by the deleted example. 

Rosetta Stone puzzles are designed to be self-contained, uniquely solvable, and accessible to ordinary solvers~\citep{WordLetterNumber}, but they are not necessarily minimal. Some puzzles rely on every context example, whereas others contain redundant evidence. Our framework exploits this distinction: deleting load-bearing examples may render part of the puzzle unsolvable, whereas deleting redundant examples preserves solvability.

\begin{figure}[t]
\centering
 \includegraphics[width=\columnwidth]{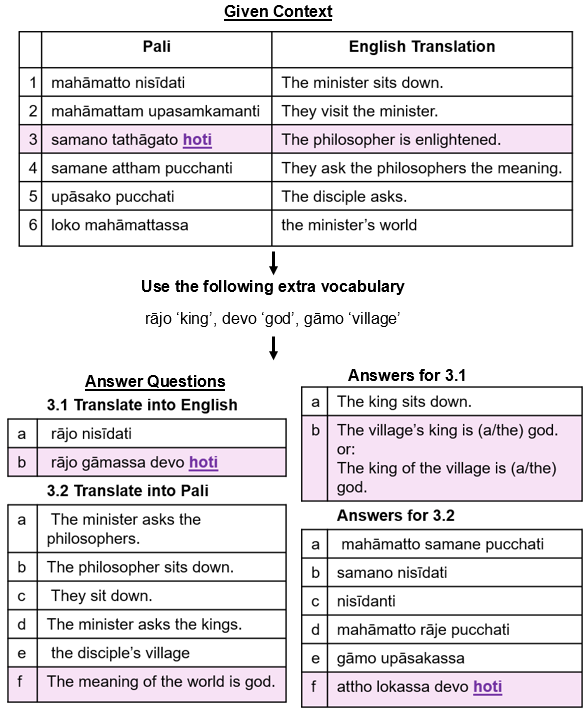}
 \caption{The Pali puzzle's context, questions, and answers (full puzzle in App.~\ref{sec:appExamples}, Fig.~\ref{fig:Pali}). Context example 3 (pink) is the only example that contains the Pali word \emph{hoti} (``is''); deleting it makes questions 3.1.b and 3.2.f (also pink) unanswerable from the remaining context.}
  \label{fig:PaliIntro}
\end{figure}

Figure~\ref{fig:PaliIntro} illustrates the deletion of a load-bearing context example. Deleting context example 3 (in pink) removes the only occurrence of the Pali word  \emph{hoti} (``is''), making questions 3.1.b and 3.2.f unanswerable from the remaining context. See App.~\ref{sec:appExamples}, Fig.~\ref{fig:Pali} for the complete Pali puzzle.

\begin{figure}[t]
\centering
 \includegraphics[width=.55\columnwidth]{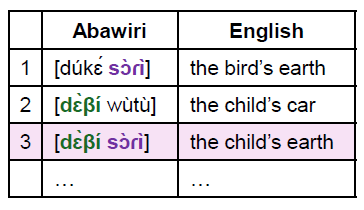}
 \caption{First three context examples of the Abawiri puzzle (full puzzle in App.~\ref{sec:appExamples}, Fig.~\ref{fig:Abawiri}).  Context example 3 (pink) is redundant: its tokens and the lexical-possession feature it demonstrates are also attested in the other two examples.}
  \label{fig:RedundantContext}
\end{figure}

In contrast, Figure~\ref{fig:RedundantContext} shows the deletion of a redundant context example. Deleting context example 3 (in pink) from the Abawiri puzzle preserves solvability as the lexical-possession and both Abawiri tokens are attested in the context elsewhere. See App.~\ref{sec:appExamples}, Fig.~\ref{fig:Abawiri} for the complete Abawiri puzzle.

For each of the 53 puzzles in the dataset containing $n$ context examples, we generate two modified versions containing $n-1$ context examples using two deletion strategies: uniform random deletion (Section~\ref{sec:random_deletion}) and ECC-inspired targeted deletion (Section~\ref{sec:ECC_deletion}).

\subsection{Uniform Random Deletion}
\label{sec:random_deletion}

The first context example deletion strategy picks a context example uniformly at random. Random deletion serves as a baseline. Removing a randomly selected context example does not necessarily affect puzzle solvability because the deleted information may be redundant. 

\subsection{ECC-inspired Targeted Deletion}
\label{sec:ECC_deletion}

The second strategy removes the context example whose deletion is expected to have the greatest structural impact on the reasoning task. To identify such examples, we introduce the \textbf{Question Damage Score}, a task-level measure that quantifies how many questions lose the information required for their solution after deleting a given context example. We motivate this strategy through an analogy to Error-Correcting Code (ECC) theory~\citep{shannon1948mathematical,hamming1950error}: context examples act as symbols and tokens as encoded information. Tokens appearing in multiple examples are redundant, whereas \emph{singleton} tokens appear only once and are lost if their source example is deleted.

We define tokens as whitespace-delimited character sequences. The Question Damage Score parameter is not intended as a complete linguistic dependency model. It is a deterministic, language-agnostic first-order approximation based on whitespace-token uniqueness. We emphasize that whitespace-token identity is a deliberately simple, first-order approximation of the linguistic dependency structure captured by the Question Damage Score, rather than a complete model of how information is represented in a puzzle. In morphologically rich languages, a whitespace-delimited word may contain a stem or morphemes that recur in other context examples, allowing information that appears unique at the word level to be partially recovered; conversely, dependencies involving morphology, agreement, word order, or interactions among examples are not explicitly represented by the current extractor. This concern is not hypothetical: Choudhary et al.~\shortcite{cunveiling} show that making morpheme boundaries explicit measurably improves LLM solvability of Linguistics Olympiad puzzles, confirming that whitespace tokens and the units models actually reason over can come apart. Our choice of whitespace tokens is therefore deliberately conservative rather than assumed to be neutral.

We adopt whitespace tokenization because it is transparent, deterministic, and uniformly applicable across our 52 languages from 34 families, many of which are low-resource and lack reliable morphological analyzers or other high-quality language-specific tools. Requiring such tools would substantially restrict the linguistic coverage that motivates this study. Accordingly, Question Damage Score should be interpreted as identifying information loss under this chosen structural representation, not as proving absolute linguistic unsolvability. The broader framework is not tied to whitespace tokenization: richer morphological or syntactic representations can replace the extraction function as suitable resources become available. 

\subsection{Formal setup}
\label{sec:secECCformalStep}
A Rosetta Stone puzzle $\mathcal{P}$ comprises three sets:
\begin{itemize}
    \item Context examples $\mathcal{L} = \{l_1, \ldots, l_n\}$, where $l_j = (f_j, e_j)$ pairs a foreign-language passage $f_j$ with its English translation $e_j$.
    \item Questions $\mathcal{Q} = \{q_1, \ldots, q_m\}$, each requiring a translation in one direction or the other.
    \item Tokens $\mathcal{F} = \{\phi_1, \ldots, \phi_k\}$, the basic informational units carried by the context examples. We define a token as a whitespace-delimited character sequence; the choice is discussed at the end of this appendix.
\end{itemize}

The distribution of tokens across context examples is captured by a binary coverage matrix $F \in \{0,1\}^{k \times n}$:
\begin{equation}
F_{ij} = \begin{cases}
1 & \text{if } l_j \text{ contains } \phi_i, \\
0 & \text{otherwise.}
\end{cases}
\end{equation}

\subsection{Damage scores}

The \emph{Base Damage Score} $D(j)$ counts the singleton tokens that context example $l_j$ uniquely provides:
\begin{equation}
D(j) = \big|\{\phi_i : F_{ij} = 1 \text{ and } \textstyle\sum_{\ell=1}^{n} F_{i\ell} = 1\}\big|.
\end{equation}
If $D(j) > 0$, deleting $l_j$ permanently removes at least $D(j)$ tokens from the puzzle's context.

To define which questions depend on a deleted example, we first identify the tokens that each question explicitly mentions. For question $q_r$, let $\tau(q_r)$ denote the question text together with its gold answer. Using the same tokenization as for $\mathcal{F}$:
\begin{equation}
\mathcal{F}_r = \{\phi_i \in \mathcal{F} : \phi_i \in \mathrm{tok}(\tau(q_r))\},
\end{equation}
where a token is counted as occurring only on a word-boundary match (so, e.g., \texttt{in} is not matched inside \texttt{into}). Thus $\mathcal{F}_r$ is the subset of puzzle tokens whose surface form appears in $\tau(q_r)$; we treat these as the tokens required to answer $q_r$ under our surface-overlap criterion.

$D(j)$ counts information loss, but not every lost token is required by a question. The \emph{Question Damage Score} $D_Q(j)$ refines this by counting only the questions made unanswerable. Let $\mathcal{F}_r \subseteq \mathcal{F}$ be the tokens required to answer question $q_r$. Then:
\begin{equation}
\resizebox{.95\hsize}{!}{$
D_Q(j) = \big|\{q_r : \exists\, \phi_i \in \mathcal{F}_r \text{ s.t.\ } F_{ij}{=}1 \text{ and } \textstyle\sum_{\ell=1}^{n} F_{i\ell}{=}1\}\big|.
$}
\end{equation}
$D_Q(j) > 0$ implies that at least one question requires a singleton token uniquely carried by $l_j$, so deleting $l_j$ makes that question unanswerable from the remaining context. This is the score the ECC-inspired strategy maximizes over $j$.

In this work, we do not normalize the $D_Q$ scores. Since the numbers of context examples and puzzle questions vary across UKLO puzzles, $D_Q$ is an absolute measure of question damage rather than a normalized measure of puzzle fragility, making direct comparisons across puzzles inappropriate. Consequently, the only meaningful cross-puzzle comparison is whether $D_Q(j)=0$, indicating that removing context example $j$ leaves the puzzle solvable under the current definition of $D_Q$.

We characterize the structural effect of deleting context example $l_j$ using complementary scores:

\begin{itemize}
\item \textbf{Base Damage Score $D(j)$}: the number of singleton tokens uniquely provided by context example $l_j$.
\item \textbf{Question Damage Score $D_Q(j)$}: the number of questions requiring singleton tokens uniquely provided by $l_j$.
\end{itemize}

$D_Q(j)>0$ indicates that deleting $l_j$ removes information required to answer at least one question.

The Question Damage Score is the operational measure used for targeted deletion. For each puzzle, we compute $D_Q(j)$ for every context example and delete the example with the largest score (ties broken arbitrarily). For example, in the Pali puzzle of Fig.~\ref{fig:PaliIntro}, the context example marked for deletion is example 3: the unique source of the word \emph{hoti}, which questions 3.1.b and 3.2.f both require.

Applying the Question Damage Score across all context examples yields a structural characterization of every puzzle:

\begin{itemize}
    \item \textbf{Robust puzzles ($\max_j D_Q(j) = 0$):} no single deletion removes question-relevant information. 7 of the 53 puzzles are robust.
    \item \textbf{Fragile puzzles ($\max_j D_Q(j) > 0$):} at least one context example is load-bearing, and ECC-targeted deletion produces a variant that is structurally unsolvable under whitespace-token reasoning. 46 of the 53 puzzles are fragile.
\end{itemize}

Structural unsolvability follows directly from our framework: under whitespace tokenization, deleting an example containing question-critical singleton tokens removes information required by at least one question. This does not imply that an LLM cannot recover the missing form through morphological or pragmatic inference, particularly in morphologically rich languages (App.~\ref{sec:appECC}). We therefore complement the structural criterion with an LLM-as-judge analysis (Section~\ref{subsec:exp2}) to distinguish clearly destructive cases from borderline ones.

The fragile/robust split is the only $D_Q$-derived quantity our results depend on (Section~\ref{sec:results}). We do not normalize $D_Q$: as the number of tokens, context examples and questions varies across UKLO puzzles, $D_Q$ is an absolute count of damaged questions rather than a fragility rate, so magnitudes are not directly comparable between puzzles. We therefore report $D_Q$ magnitudes descriptively and base all cross-puzzle classification on whether $D_Q(j)=0$. Our current implementation adopts an approximation based on whitespace tokenization and question-token dependencies, leaving richer structural relationships for future work (e.g., token order, proximity, interactions, other dependencies).

\subsection{Scope and Generality}

$D_Q$ is defined over an abstract coverage matrix $F$ and question requirement sets $F_r$ (Sec.~\ref{sec:secECCformalStep}); only the extraction function that populates $F$ (whitespace-token identity) is task-specific. The framework applies to any task that is (i) self-contained, (ii) modular under single-unit removal, (iii) equipped with a computable unit–question dependency relation, and (iv) accompanied by a verified answer key. Condition (i) binds: where prior knowledge overlaps task content, deletion shifts derivation to parametric memory rather than removing it, and structural damage becomes uninterpretable. Rosetta Stone puzzles satisfy all four by construction. Multi-hop QA with annotated supporting sentences and math word problems with explicit relational structure satisfy them under stronger supervision, with $D_Q$ unchanged and only the extractor differing. We validate empirically only on linguistic puzzles here.

\section{Experimental Design}
\label{sec:experiment}
\label{sec:experiment}

We conduct two experiments. The first experiment evaluates whether models solve the puzzles and abstain when the available context is structurally insufficient. The second one uses LLMs as judges to provide an independent, language-aware assessment of whether modified puzzles remain solvable.

We evaluate three frontier models: Claude Sonnet 4.6~\citep{anthropic2026claudesonnet46}, Gemini 3 Flash~\citep{google2026gemini3flash}, and GPT-5.4~\citep{openai2026gpt54}. All models are queried via their APIs using default settings in a single zero-shot pass per puzzle. See App.~\ref{sec:appPrompts} for model versions and access dates.

\subsection{Puzzle Solving under Controlled Context Perturbations}
\label{subsec:exp1}

We use a zero-shot structured-output setup to evaluate each model on the original puzzles and two modified versions: one with a randomly deleted context example and one with an ECC-inspired targeted deletion, yielding 159 puzzle instances. Each puzzle is presented as a single prompt containing the preamble, context examples, and questions. The system prompt instructs the model to rely exclusively on the provided context and output \texttt{N/A} whenever a question cannot be answered. The full prompt is provided in App.~\ref{subsec:appPromptsExp1}.

Performance is measured using Exact Match (EM) against the official UKLO answers. For the modified puzzles, we additionally record abstentions and compare performance across the original, random-deletion, and targeted-deletion conditions.

\subsection{Structural vs Model-Perceived Solvability}
\label{subsec:exp2}

The Question Damage Score provides a deterministic structural characterization of each modified puzzle under the assumptions described in Section~\ref{sec:ECC_deletion}. Specifically, deleting a context example with $D_Q(j)>0$ removes information required to answer at least one question. However, structural analysis alone cannot determine whether the remaining information still permits the original answers through alternative reasoning strategies, redundancy, or prior linguistic knowledge.

Expert adjudication of solvability is confounded rather than merely costly. Determining whether the original answers remain derivable from the reduced context is itself a reasoning problem, and because the puzzles are self-contained by design, a specialist who already knows the morphology cannot separate puzzle-internal derivability from their own knowledge of the language. Scale compounds this across 106 variants in 53 puzzles from 52 languages (34 language families). 

We therefore employ LLM-as-a-Judge as an auxiliary semantic validation of the structural criterion rather than as its replacement~\citep{zheng2023judging}. Each judge model receives the original puzzle together with its answer key and a modified puzzle, and determines whether the original answers remain derivable from the modified context. The complete prompt is provided in App.~\ref{subsec:appPromptsExp2}.

The judge's assessment is interpreted independently of the structural analysis. LLM-as-a-Judge's agreement strengthens confidence in the structural criterion, while disagreements highlight the distinction between the logical structure of the reasoning task and model behavior.

\subsection{Complete Context Removal}

We also compare our framework with complete context removal, a setting commonly used to study LLM performance on linguistic puzzles~\citep{LingOly,khouja2026lingolytoodisentanglingreasoningknowledge}. 

\section{Results}
\label{sec:results}
\label{sec:results}

\begin{table*}[t]
\centering
\small
\begin{tabular}{@{} l c c c c c c c c c @{}}
\toprule
& \multicolumn{3}{c}{\textbf{Original}} & \multicolumn{3}{c}{\textbf{Random}} & \multicolumn{3}{c}{\textbf{ECC-inspired}} \\
\cmidrule(lr){2-4} \cmidrule(lr){5-7} \cmidrule(lr){8-10}
\textbf{Model} & \textbf{w/ N/A} & \textbf{\#N/A} & \textbf{w/o N/A} & \textbf{w/ N/A} & \textbf{\#N/A} & \textbf{w/o N/A} & \textbf{w/ N/A} & \textbf{\#N/A} & \textbf{w/o N/A} \\
\midrule
GPT-5.4            & 0.425 & 1  & 0.433 & 0.407 & 1 (0)  & 0.415 & 0.406 & 1 (0)  & 0.415 \\
Claude Sonnet 4.6  & 0.313 & 13 & 0.397 & 0.241 & 15 (10) & 0.312 & 0.293 & 15 (11) & 0.311 \\
Gemini 3 Flash     & 0.730 & 3  & 0.747 & 0.697 & 2 (0)  & 0.702 & 0.662 & 4 (2) & 0.704 \\
\bottomrule
\end{tabular}
\caption{Average Exact Match (EM) scores across the three models and three puzzle versions.  \textbf{w/ N/A} treats abstentions as score 0; \textbf{\#N/A} counts puzzles where the model abstained or produced no output; \textbf{w/o N/A} averages over only the puzzles the model attempted. In the two deletion conditions, the parenthesized figure is the deletion-driven subset of \textbf{\#N/A} — puzzles the model abstained on after deletion but answered in the original condition. With 53 puzzles per condition, small aggregate gaps are not informative on their own; the per-puzzle analysis below is what carries the diagnostic load.}
\label{tab:exact_match_scores}
\end{table*}


Our evaluation combines three complementary perspectives on modified puzzles. \textbf{The Question Damage Score} ($D_Q$) provides a model-agnostic structural criterion for identifying deletions that remove information required by the intended puzzle solution. The \textbf{LLM-as-a-judge experiment} provides an independent semantic assessment of whether a modified puzzle remains solvable. Finally, the \textbf{solver experiment} measures whether a model abstains or continues producing answers under the modified context. Together, these experiments allow us to distinguish structural sufficiency, model-perceived solvability, and actual model behavior.

\subsection{Puzzle Solving and Abstention}
\label{ref:sebsecPuzzleSolvingAbstention}

We begin by comparing our framework with complete context removal, which has been used in prior studies of linguistic reasoning~\citep{LingOly,khouja2026lingolytoodisentanglingreasoningknowledge}. Removing the entire context eliminates the intended reasoning task, leaving memorized knowledge as the only possible source of correct answers. Consistent with prior work, all three frontier LLMs rarely abstain and occasionally answer questions correctly despite receiving no puzzle context (See App.~\ref{sec:appNoContext}). While this baseline demonstrates that models continue producing answers in the absence of supporting evidence, it does not distinguish between memorization, inference, or other sources of success.

In contrast, our framework preserves the overall reasoning task while removing a single load-bearing context example. This allows us to identify cases where structural sufficiency, model-perceived solvability, and solver behavior diverge, providing a controlled setting for future investigations of the mechanisms underlying these discrepancies.

Table~\ref{tab:exact_match_scores} summarizes the performance across the original puzzles and two single-example deletion strategies. Aggregate Exact Match (EM) changes only modestly after deletion, largely because many deleted examples are redundant. Moreover, models almost never abstain despite being instructed to do so when the available context is insufficient.

Gemini~3~Flash performs best on original puzzles without N/A (0.747), followed by GPT-5.4 (0.433) and Claude Sonnet~4.6 (0.397). For reference, the UKLO human baseline on these 53 puzzles, computed from the highest reported difficulty tier for each puzzle, is 0.484. All models show small decreases on modified variants while attempting nearly the same number of puzzles. Given $N=53$, these aggregate differences fall within expected single-sample variability, except for Gemini's significant drop.

These aggregate numbers understate the effect by construction. ECC-targeted deletion damages $D_Q(j)$ of a puzzle's $m$ questions, capping the possible EM drop at $D_Q(j)/m$, so puzzle-level EM averages damaged questions with an undamaged majority and aggregate stability is expected either way. We therefore condition on structural labels and judge verdicts (Section~\ref{subsec:Contractions}) rather than on condition-level means.

Across the 106 modified puzzles, GPT-5.4 abstains twice, Gemini six times, and Claude thirty times (Table~\ref{tab:exact_match_scores}). Most of these are not responses to the deletion. GPT abstains on a single puzzle, and does so identically on the original, so no abstention is attributable to a deletion. Gemini's six reduce to two, both under ECC-targeted deletion; Claude's thirty reduce to twenty-one, split almost evenly between random (10) and targeted (11) deletion. Once pre-existing abstentions are excluded, only Gemini's deletion-driven abstentions occur exclusively under targeted deletion, and they amount to two puzzles; Claude's split evenly (10 vs. 11). Abstention does not track structural damage even for the model that abstains most.


\subsection{Structural Criterion versus LLM-as-a-Judge}
\label{ref:seubsection:StructuralCriterion}

The rarity of abstentions indicates that even after the removal of question-critical context, models overwhelmingly prefer to produce an answer rather than acknowledge insufficient information. This pattern motivates the subsequent analyses, which distinguish whether the modified puzzles remain structurally sufficient ($D_Q$), appear solvable to the models themselves (LLM-as-a-judge), and how the models ultimately behave as solvers.

The structural criterion and LLM-as-a-judge measure different notions of solvability. While $D_Q$ identifies deletions that remove information required under the intended puzzle reasoning process, LLM-as-a-judge evaluates whether the modified puzzle still appears solvable given the model's own linguistic knowledge and inference abilities.

Table~\ref{tab:judge_alignment_metrics} compares each model's judge verdict on the ECC-modified puzzles against the structural classification ($\max D_Q > 0 \Rightarrow$ fragile $\Rightarrow$ unsolvable; $\max D_Q = 0 \Rightarrow$ robust). Claude is the most conservative judge: it never labels a robust puzzle unsolvable (100\% precision) but misses 18 of 46 fragile puzzles (60.9\% recall). Gemini is the strongest overall agreement: 33 of 46 fragile puzzles correctly labeled (71.7\% recall) at 97.1\% precision.
GPT is the weakest, identifying half of the fragile puzzles as unsolvable.
We would like to emphasize that (1) The LLM judge is an auxiliary cross-check, rather than the ground truth; (2) Human expert adjudication is not feasible at this scale: the dataset spans 52 languages from 34 families, many low-resource, making it impractical to recruit qualified experts for every language and every modified puzzle variant; moreover, experts’ prior linguistic knowledge could itself confound judgments of whether an answer is derivable strictly from the remaining puzzle context.

\begin{table}[t]
\centering \small
\begin{tabular}{l c c c c c c}
\toprule
& \multicolumn{2}{c}{\textbf{Gemini}} & \multicolumn{2}{c}{\textbf{GPT}} & \multicolumn{2}{c}{\textbf{Sonnet}}  \\
\cmidrule(lr){2-3} \cmidrule(lr){4-5} \cmidrule(lr){6-7}
\textbf{Fragility} & \textbf{U.} & \textbf{V.} & \textbf{U.} & \textbf{V.} & \textbf{U.} & \textbf{V.} \\
\midrule
\textbf{Fragile} ($\max D_Q > 0$) & 33 & 13 & 23 & 23 & 28 & 18 \\
\textbf{Robust}  ($\max D_Q = 0$) & 1  & 6  & 0  & 7  & 0  & 7  \\
\midrule
\textbf{Total}   & 34 & 19 & 23 & 30 & 28 & 25 \\
\midrule
\multicolumn{7}{l}{\textbf{Detecting `Unsolvable' (positive class)}} \\
\midrule
\textbf{Precision} & \multicolumn{2}{c}{97.1\%} & \multicolumn{2}{c}{\textbf{100.0\%}} & \multicolumn{2}{c}{\textbf{100.0\%}} \\
\textbf{Recall}  & \multicolumn{2}{c}{\textbf{71.7\%}} & \multicolumn{2}{c}{50.0\%} & \multicolumn{2}{c}{60.9\%} \\
\bottomrule
\end{tabular}
\caption{Each model's judgments on the ECC-modified puzzles, compared against the structural classification by $\max D_Q$. U. = Unsolvable, V. = Valid. Positive class: ``Unsolvable.''}
\label{tab:judge_alignment_metrics}
\end{table}

\subsection{Contradictions Analysis}
\label{subsec:Contractions}

Our diagnostic framework enables analyses that are impossible with complete context removal. As a first example, we examine cases where LLM-as-a-judge marks a modified puzzle to be unsolvable yet subsequently answers it correctly.

Using Gemini 3 Flash, the strongest solver and the judge with the highest agreement with the structural criterion, we identify \textbf{21 of 53} puzzles in which performance on a judge-confirmed unsolvable variant matches or exceeds performance on the original puzzle. In \textbf{7} puzzles, performance strictly improves. The full breakdown of these anomalous outcomes is cataloged in Table~\ref{tab:combined_contamination}. These cases isolate instances where models produce correct answers after the derivation path is structurally severed under whitespace-token reasoning, an outcome that context-based deduction alone does not account for. Table~\ref{tab:combined_derailment} represents the converse situation. It captures cases where the puzzle remains structurally solvable after a redundant deletion, yet the model's performance strictly deteriorates. 
Together, these two tables illustrate a two-sided brittleness: the models frequently fail to exploit sufficient context, while at the same time producing correct answers that the remaining context cannot support.

Among the cases of strictly improved performance on modified puzzles, the most striking example is the Cherokee puzzle: Gemini's EM rises from 0.700 on the original puzzle to 0.900 after ECC-targeted deletion of an example that Gemini itself judges as containing critical information. The Karelian puzzle exhibits the same behavior under random deletion, going from 0.310 on the original to 0.540 on a variant judged as unsolvable.

The seven strict-improvement cases are especially difficult to reconcile with context-based deduction: removing information should not improve performance if the model relies on the modified context to derive its answers. The remaining 14 ``equals-or-exceeds'' cases have $\Delta=0$, meaning that the model produces identical outputs regardless of whether the critical context example is present. This behavior suggests that, for these puzzles, the modified context has little effect on the model's predictions. One possible explanation is recall from pretraining: if a puzzle or closely related linguistic pattern was encountered during training, performance may become partially independent of the provided context. Alternative explanations are also possible, including inference from latent linguistic regularities not captured by our structural analysis. App.~\ref{sec:appResultsTables}, Table~\ref{tab:combined_verdicts_performance} reports per-puzzle judge verdicts (with the seven strict-improvement cases highlighted), and Table~\ref{tab:combined_contamination} lists all 21 cases where performance on a judged-unsolvable variant matches or exceeds the original. The reverse pattern also occurs: models sometimes fail on puzzles judged solvable. Table~\ref{tab:combined_derailment} shows that deleting a redundant example can reduce EM by as much as 0.4 while preserving solvability.

The seven strict-improvement cases and the 14 ties are supported by two independent arguments. The first is structural: for fragile puzzles ($\max D_Q > 0$), deleting a context example removes a singleton token explicitly required by at least one question. This criterion is deterministic and independent of any model. The second is empirical: Gemini, given the original puzzle and its answer key, independently judges that the modified context is insufficient to derive the original answers. The highlighted puzzles satisfy both criteria.

We do not claim formal unsolvability in a proof-theoretic sense; a sufficiently strong inference procedure could recover some deleted information through morphological or contextual reasoning. We claim structural unsolvability under whitespace-token reasoning, corroborated by an independent language-aware judge. 

There is a clear contrast between the aggregate and puzzle-level views. Table~\ref{tab:exact_match_scores} shows only small EM differences between original and ECC-modified runs (1.8--8.6 w/o-N/A EM points across models), which on a 53-puzzle corpus are within expected sampling variability. Conditioning on structural labels ($\max D_Q$) and judge verdicts isolates a subset of independently verified destructive deletions, where the contradiction becomes pronounced: Cherokee improves by 20 EM points after deletion of critical information, and Karelian improves by 23 points under random deletion. This illustrates the main advantage of the framework: aggregate statistics may obscure effects that become visible at the level of individual context examples.

We also perform paired Wilcoxon signed-rank tests~\citep{wilcoxon1945} on puzzle-level EM scores and McNemar's tests~\citep{mcnemar1947note} on question-level outcomes (App.~\ref{sec:appStatSign}).

\section{Discussion and Conclusion}
\label{sec:discussion}

Targeted context deletion exposes systematic divergences between structural sufficiency, model-perceived solvability, and solver behavior. Frontier LLMs rarely abstain after load-bearing context is removed, and on judge-confirmed damaged variants they perform as well as or better than on the originals. The presented framework does not adjudicate whether these divergences arise from memorization, prior linguistic knowledge, or inference beyond the intended puzzle structure; it provides a controlled setting in which those explanations can be separated by targeted follow-ups.

The longer-term value of our work is interpretability: each puzzle becomes a small graph of context-example→question dependencies that can be probed one node at a time. Extending to multi-example deletion, analogous to stopping-set analysis in LDPC codes~\citep{gallager1962}, would surface interactions among examples, and the $D_Q$ annotations identify candidate puzzles for targeted contamination studies~\citep{chi-etal-2024-modeling}. We release the 53-puzzle corpus with per-example $D_Q$ scores and structural annotations; its value lies in annotation depth rather than scale.


\section*{Limitations}
\label{sec:limitations}

The goal of this work is diagnostic, not causal: the framework identifies cases where model behavior diverges from the information available under the structural criterion; it does not claim to distinguish memorization, contamination, prior linguistic knowledge, or alternative inference. 

\paragraph{Corpus scope.} The empirical study covers 53 puzzles, three frontier models, and a single random + single ECC variant per puzzle. With $n = 53$, aggregate Exact Match differences between conditions are small and would be sensitive to additional random draws if read as point estimates. We do not lean on aggregate gaps in our conclusions; the contradiction signal we report is conditioned on (i) the structural fragility classification, which does not depend on a random draw, and (ii) the per-puzzle judge verdict. A larger random-deletion ensemble (multiple random draws per puzzle) would tighten the random-deletion baseline and is a natural extension; the structural ECC argument is by construction deterministic.

\paragraph{Whitespace tokenization.} Our damage score uses whitespace tokens, which is a conservative lower bound. A whitespace-singleton may decompose into morphemes that recur elsewhere in the puzzle, particularly in agglutinative languages (App.~\ref{sec:appECC}). The structural unsolvability argument remains sound under any finer segmentation, but the whitespace bound may classify some morpheme-recoverable puzzles as fragile. The LLM-as-judge cross-check is designed to catch exactly these cases, but it is itself an LLM and is not a substitute for expert human annotation.


\paragraph{Single-example deletion.} We delete one context example at a time. Many UKLO puzzles require composing information from a chain of examples, and the framework in its current form does not surface those composition structures. Extending to multi-example deletion, formally to the stopping-set analysis of the underlying ECC, is a direction the framework is built to support and that we discuss in Section~\ref{sec:discussion}.

\paragraph{Memorization vs. contamination vs. heuristics.} The contradiction pattern we surface is incompatible with derivation from the modified context, but it admits several non-derivation explanations: literal memorization of the published UKLO answer; pattern completion via high-resource-language priors; or some combination. We do not separate these explanations in this paper; the value of the framework is to surface the candidate set of contradiction cases on a per-puzzle basis so that targeted follow-ups (paraphrase robustness, freshly authored puzzles in extinct languages, attribution-based probes) can disentangle them.

\paragraph{Model selection.} We evaluate three frontier models. Reasoning-mode ablations (e.g., extended-thinking variants) and additional models, including larger Gemini variants, are out of scope here; the same dataset and protocol apply unchanged, and the structural fragility labels do not depend on the choice of solver.

\section*{Data and Code Availability}
\label{sec:DataAndCode}
The corpus and all structural annotations are available publicly.\footnote{https://github.com/Lewy09-Tm25/LLM-Abstention-Ling-Puzzles} The release comprises the 53 UKLO Rosetta Stone puzzles in structured JSON, the random-deletion and ECC-targeted variants of each puzzle, per-context-example Base Damage Scores $D(j)$ and Question Damage Scores $D_Q(j)$, fragility labels, judge verdicts from all three models, and the prompts and evaluation scripts used to produce the reported results. The underlying puzzles are published by the UK Linguistics Olympiad and remain the property of their respective authors; we release our annotations and derived variants, and link to the original UKLO sources. The repository will be deanonymized upon acceptance.

\bibliography{custom}
\clearpage

\appendix

\section{UKLO Linguistic Puzzle Examples}
\label{sec:appExamples}

\begin{figure}[htbp]
\centering
 \textbf{Rosetta Stone Format Example \\
 Pali Puzzle:} UKLO, 2013
 \includegraphics[width=\columnwidth]{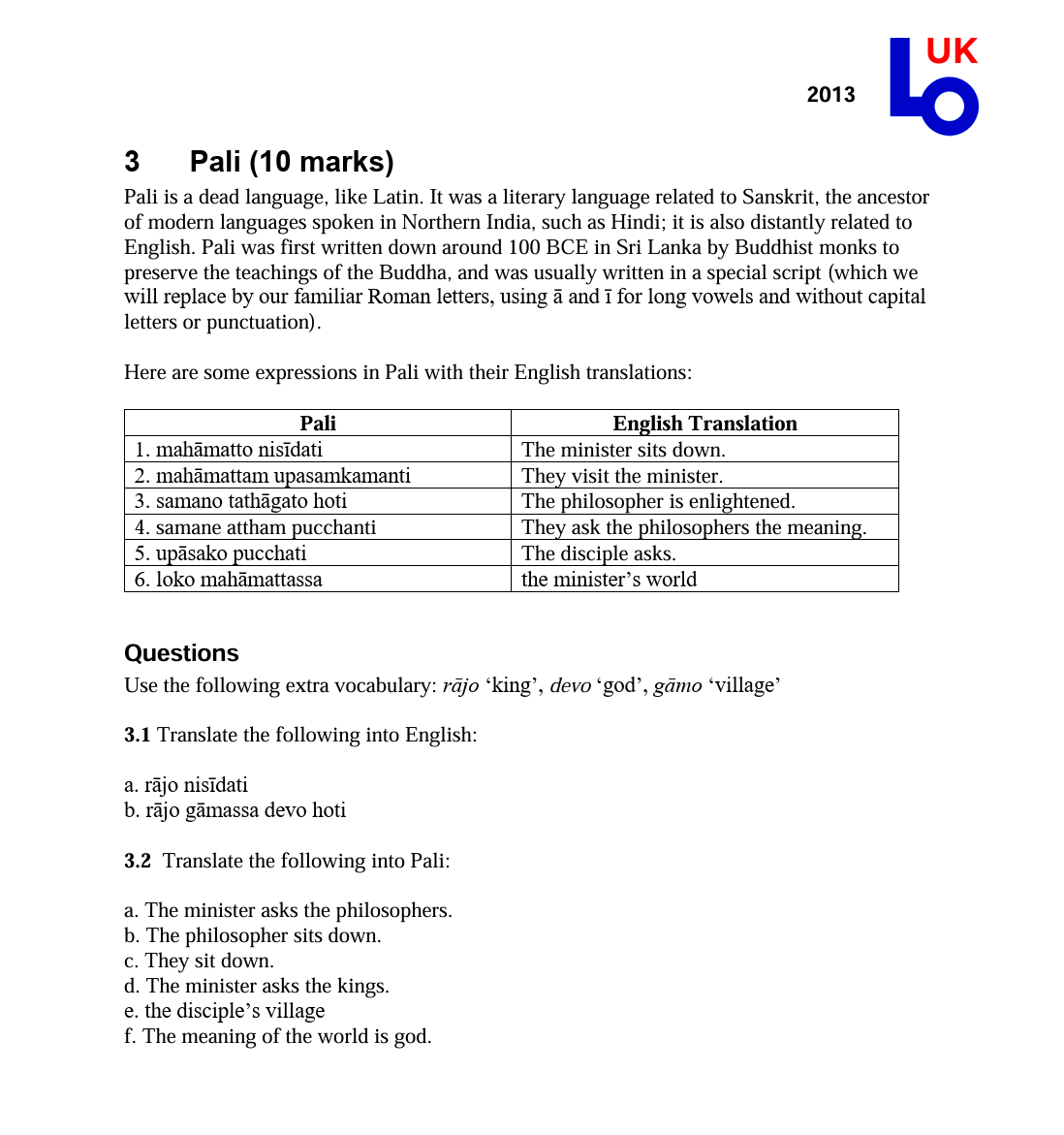}
 \caption{The Pali puzzle was used in UKLO in 2013. This puzzle has two difficulty scores: its score for the Foundation participants is $39\%$ and its score for the Intermediate participants $61\%$; its linguistic topic is a combination of Morphology and Syntax; its format is Rosetta; its language family is Indo-European, Indo-Aryan; its Author is Babette Verhoeven. \\ \url{https://www.uklo.org/wp-content/uploads/2022/09/2013.3-Pali.pdf}}  
  \label{fig:Pali}
\end{figure}

\begin{figure}[htbp]
\centering
 \textbf{Puzzle with Redundancy in Context Examples\\
 Abawiri Puzzle:} UKLO, 2023
 \includegraphics[width=\columnwidth]{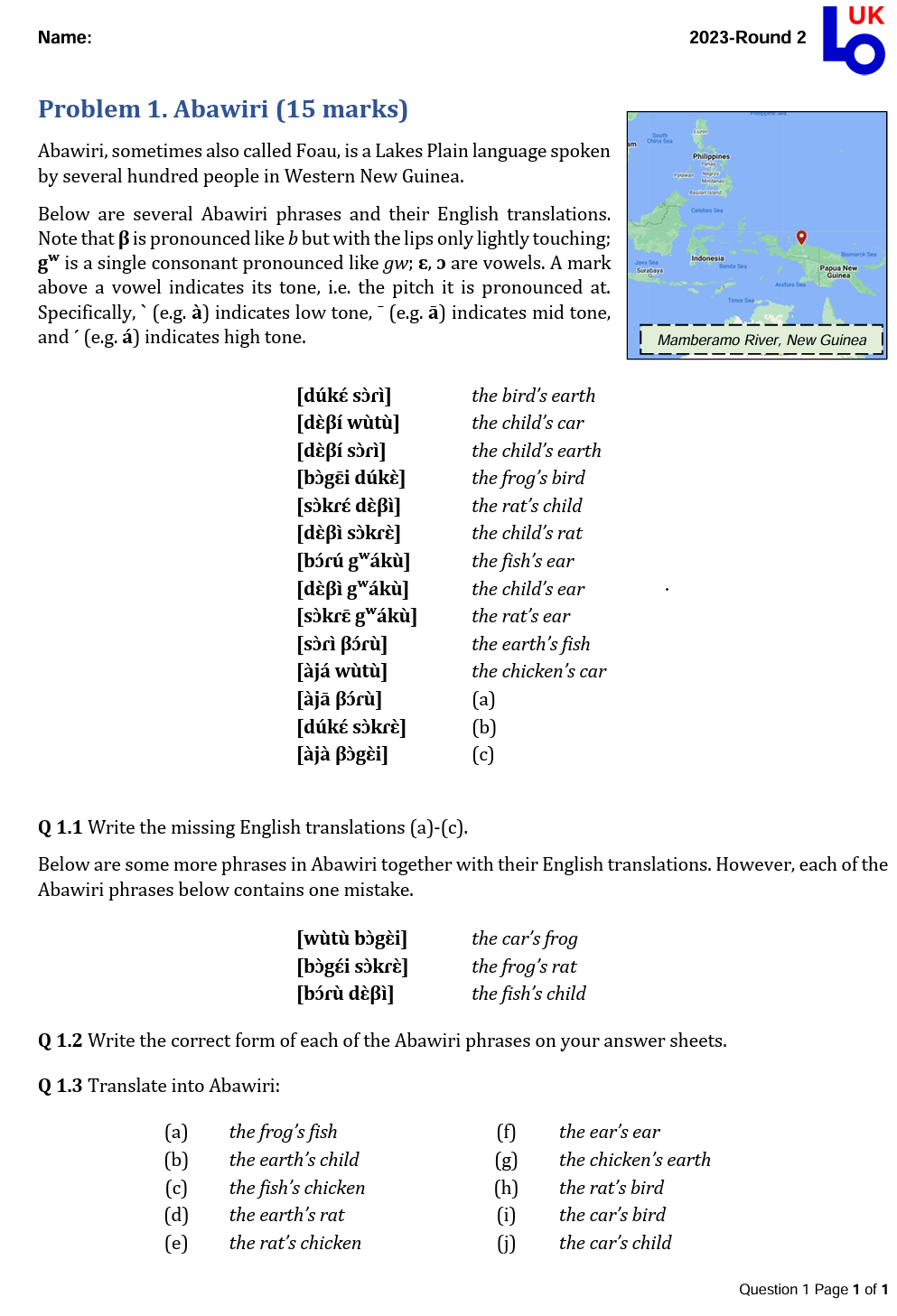}
 \caption{The Abawiri Puzzle was used in UKLO in 2023. This puzzle has one difficulty score: its score for the Round 2 participants is $60\%$; its linguistic topic is Phonology; its format is Rosetta; its language family is Lakes Plain; its Author is Daniel Lovsted. \\ \url{https://www.uklo.org/wp-content/uploads/2023/03/2023_R2_1-Abawiri.pdf}}  
  \label{fig:Abawiri}
\end{figure}

\begin{figure}[htbp]
\centering
 \textbf{Examples of Tokens with Different Linguistic Functions\\
 Permyak Puzzle:} UKLO, 2023
 \includegraphics[width=\columnwidth]{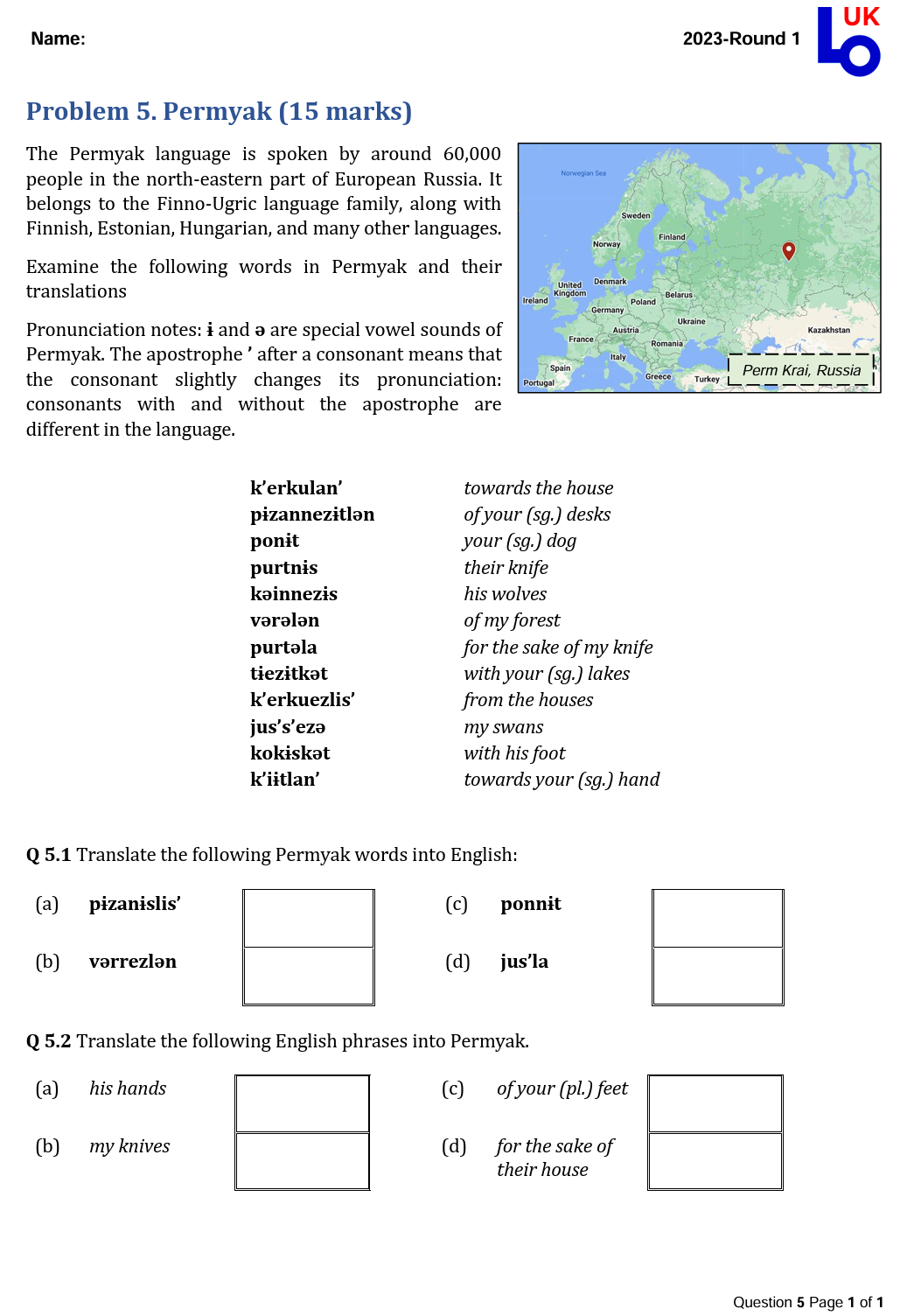}
 \caption{The Permyak puzzle was used in UKLO in 2023. This puzzle has two difficulty scores: its score for the Foundation  participants is $25\%$ and its score for the Intermediate participants $44\%$; its linguistic topic is Morphology; its format is Rosetta; its language family is Uralic; its Author is Pavel Iosad. \\ \url{https://www.uklo.org/wp-content/uploads/2023/03/2023_R1_5-Permyak.pdf}}  
  \label{fig:Permyak}
\end{figure}

\begin{figure}[htbp]
\centering
 \textbf{Rosetta Stone Example with Pictorial Context\\
 Chinese Characters Puzzle:} UKLO, 2016
 \includegraphics[width=0.8\columnwidth]{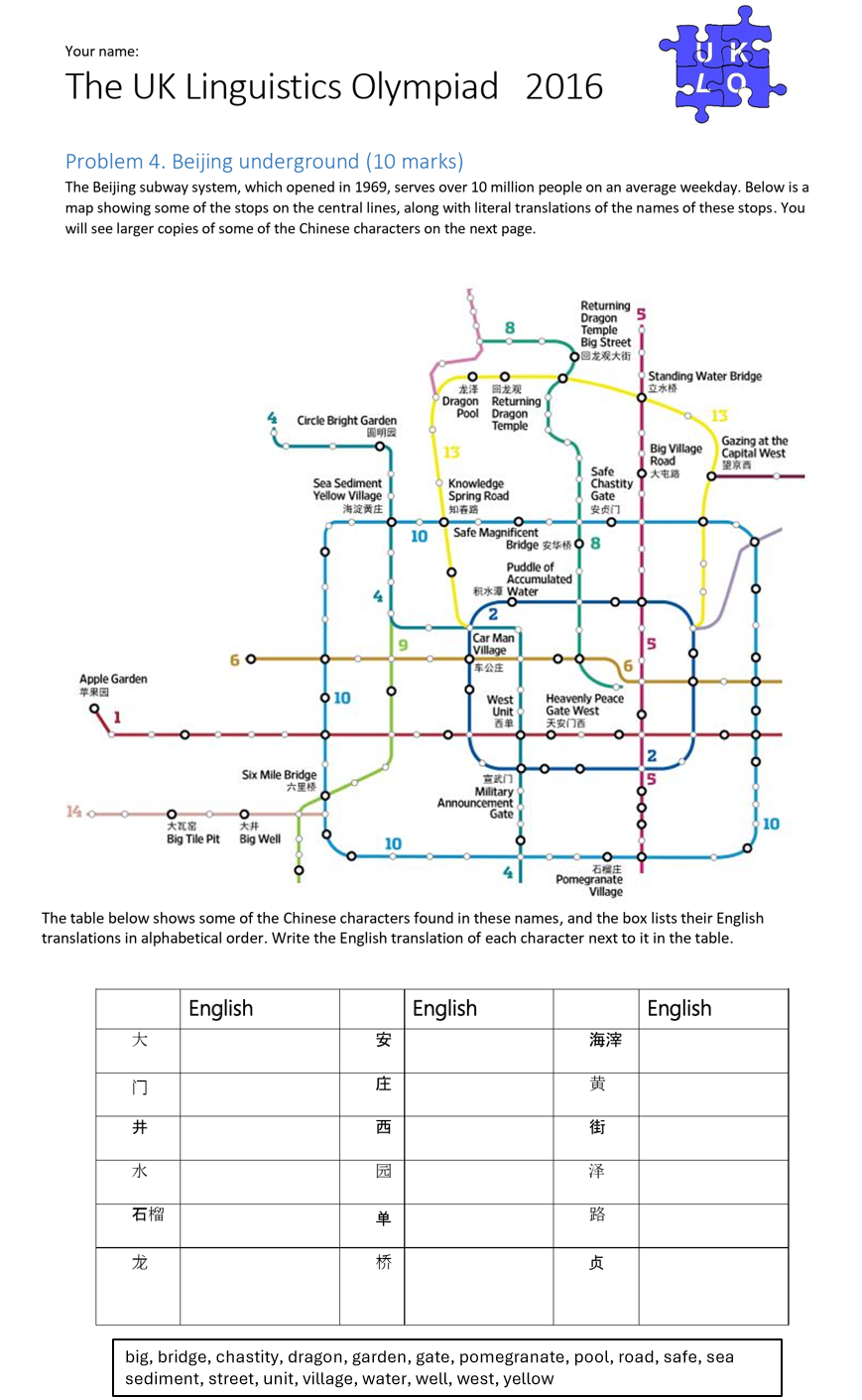}
 \caption{The Chinese Characters puzzle was used in UKLO in 2016. This puzzle has two difficulty scores: its score for the Foundation participants is $74\%$ and its score for the Intermediate participants $92\%$; its linguistic topic is Writing System; its format is Rosetta; its language family is Sino-Tibetan; its Author is Catherine Sheard. \\ \url{https://www.uklo.org/wp-content/uploads/2022/05/2016_4.-Beijing.pdf}}  
  \label{fig:Chinese}
\end{figure}

\begin{figure}[htbp]
\centering
 \textbf{Rosetta Stone Example with Multiple Perspectives\\
 Longgu Puzzle:} UKLO, 2021
 \includegraphics[width=\columnwidth]{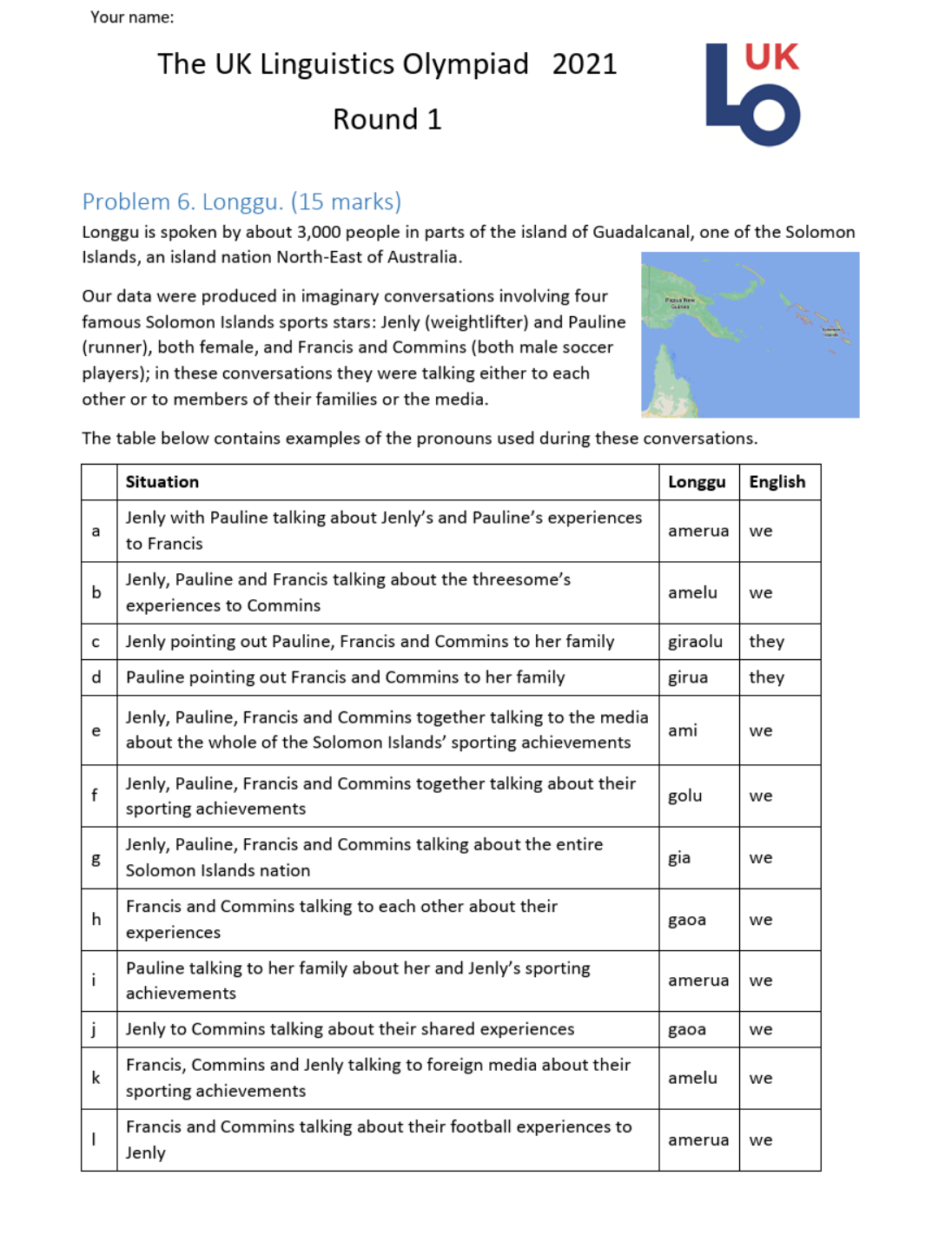}
 \caption{The Longgu puzzle was used in UKLO in 2021. This puzzle has one difficulty score: its score for the Intermediate participants is $37\%$; its linguistic topic is Semantics; its format is Rosetta; its language family is Austronesian, Oceanic; its Author is Babette Verhoeven. \\ \url{https://www.uklo.org/wp-content/uploads/2022/05/2021_6-Longgu.pdf}}  
  \label{fig:Longgu}
\end{figure}

\begin{figure}[htbp]
\centering
 \textbf{Rosetta Stone Example with More Than Two Languages\\
 Korowai and Haruai Puzzle:} UKLO, 2022
 \includegraphics[width=\columnwidth]{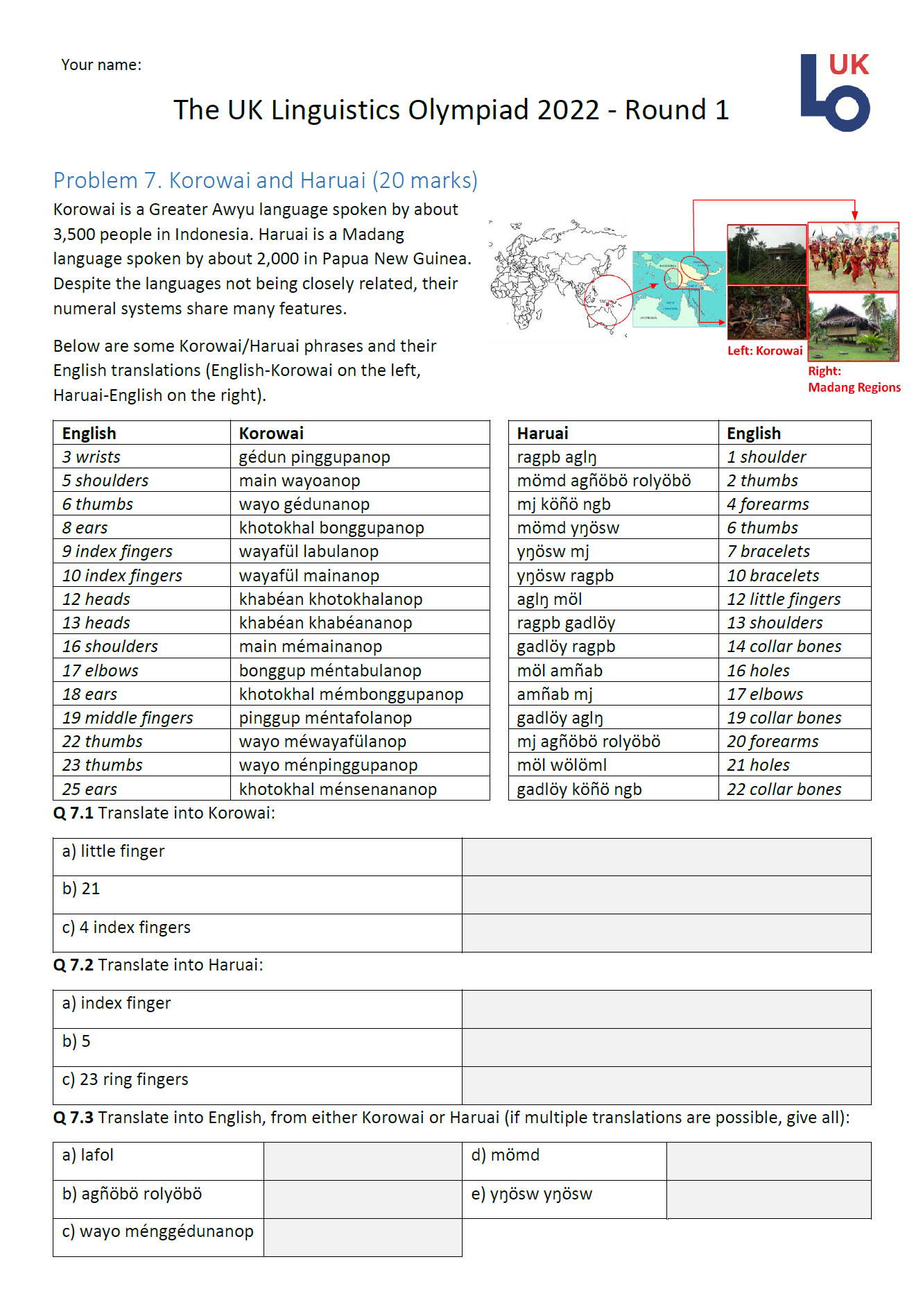}
 \caption{The Korowai and Haruai puzzle was used in UKLO in 2022. This puzzle has two difficulty scores: its score for the Intermediate participants is $22\%$ and its score for the Advanced participants $26\%$; its linguistic topic is Numbers; its format is Rosetta; its language family is Multiple (Trans-New-Guinea, Piawi); its Author is Simi Hellsten. \\ \url{https://www.uklo.org/wp-content/uploads/2022/08/7_UKLO-2022-Korowai-and-Haruai__Complete-Script.zip}}  
  \label{fig:Korowai}
\end{figure}

\begin{figure}[htbp]
\centering
 \textbf{Match-Up Format Example \\
 Waama Puzzle:} UKLO, 2021
 \includegraphics[width=\columnwidth]{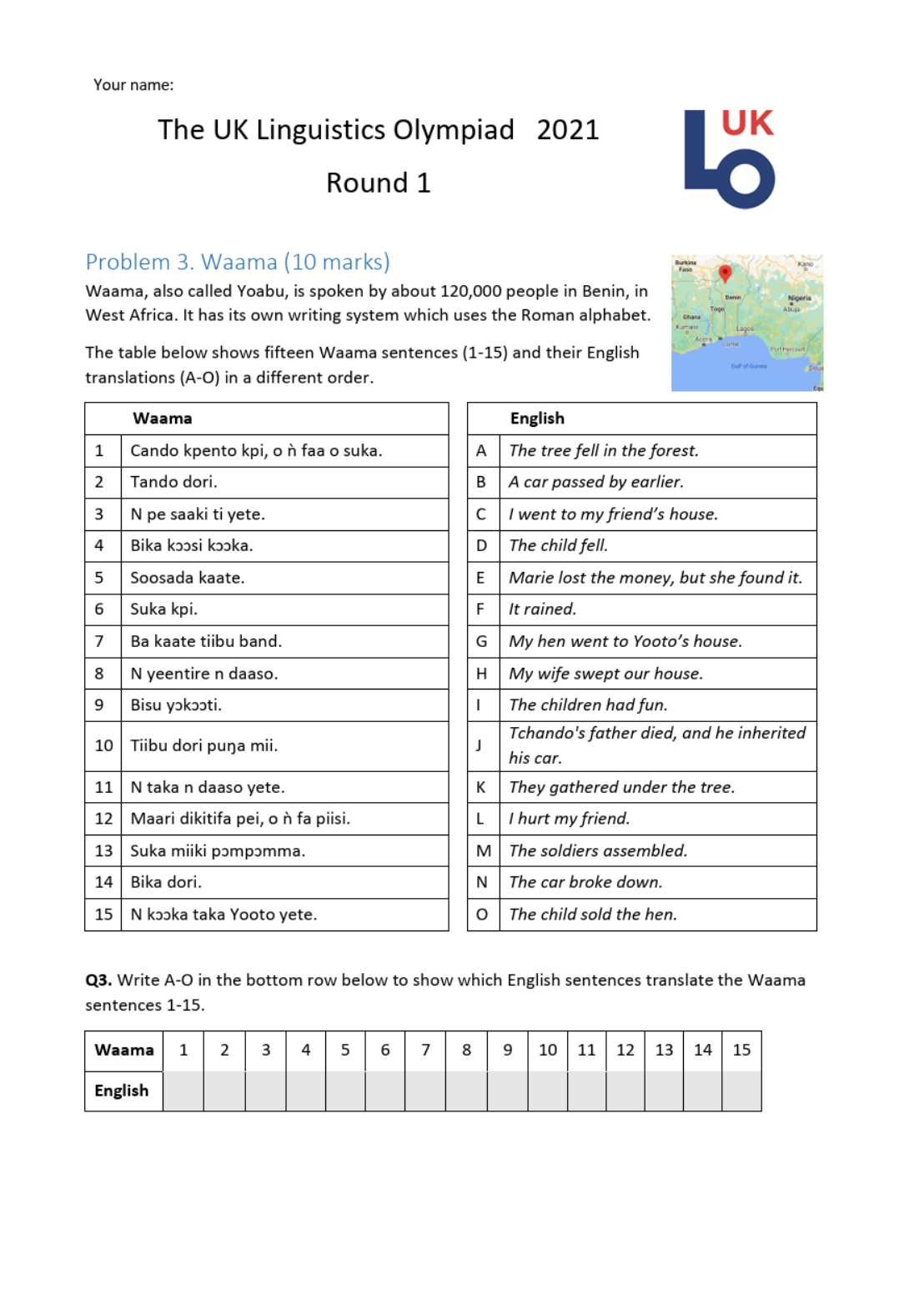}
 \caption{The Waama puzzle was used in UKLO in 2021. This puzzle has two difficulty scores: its score for the Breakthrough participants is $42\%$ and its score for the Foundation participants $54\%$; its linguistic topic is Syntax; its format is Match-Up; its language family is Atlantic-Congo, Gur; its Author is Aleka Blackwell. \\ \url{https://www.uklo.org/wp-content/uploads/2022/05/2021_3-Waama.pdf}}  
  \label{fig:Waama}
\end{figure}
\clearpage

\section{Context Deletion Examples}
\label{sec:appTreeCut}
\begin{figure}[h!]
\centering
 \includegraphics[width=.99\columnwidth]{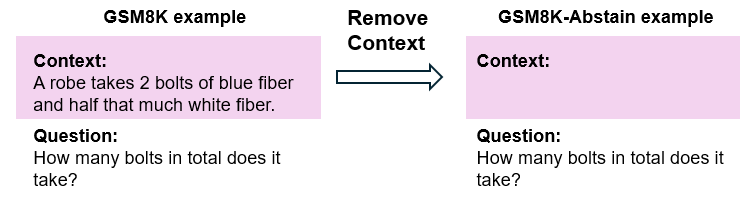}
 \caption{Context removal in \citet{AbstentionBench}: the original GSM8K problem statement is stripped, leaving a question with no derivable answer.}
  \label{fig:ContextRemoval}
\end{figure}

Several recent approaches generate unanswerable problems by removing information rather than constructing adversarial text. GSM8K-Abstain~\citep{AbstentionBench}, for example, removes the entire problem context, making each resulting instance trivially unanswerable (Fig.~\ref{fig:ContextRemoval}).

\begin{figure}[htbp]
\centering
 \textbf{TreeCut Example} \\
 \includegraphics[width=\columnwidth]{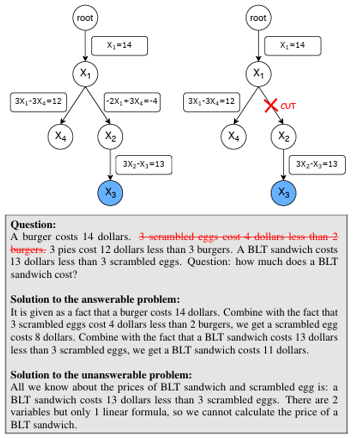}
 \caption{The TreeCut deletion example from~\citep{ouyang-2025-treecut}}  
  \label{fig:TreeCut}
\end{figure}

The example in Fig.~\ref{fig:TreeCut} corresponds to the Figure 1 example from~\citep{ouyang-2025-treecut}. In Fig.~\ref{fig:TreeCut}, the top panels depict the tree structures corresponding to the answerable and unanswerable questions, respectively. In the bottom panel, the strike-through sentence represents the formula removed by the cut. The variable mappings to items are as follows: $x_1$ represents a burger, $x_2$ represents a scrambled egg, $x_3$ represents a BLT sandwich, and $x_4$ represents a pie.

\section{ECC-Inspired Damage Score and Whitespace Tokanization}
\label{sec:appECC}

\begin{figure*}[t]
\centering
 \textbf{Permyak puzzle:} UKLO, 2023
 \includegraphics[width=.75\textwidth]{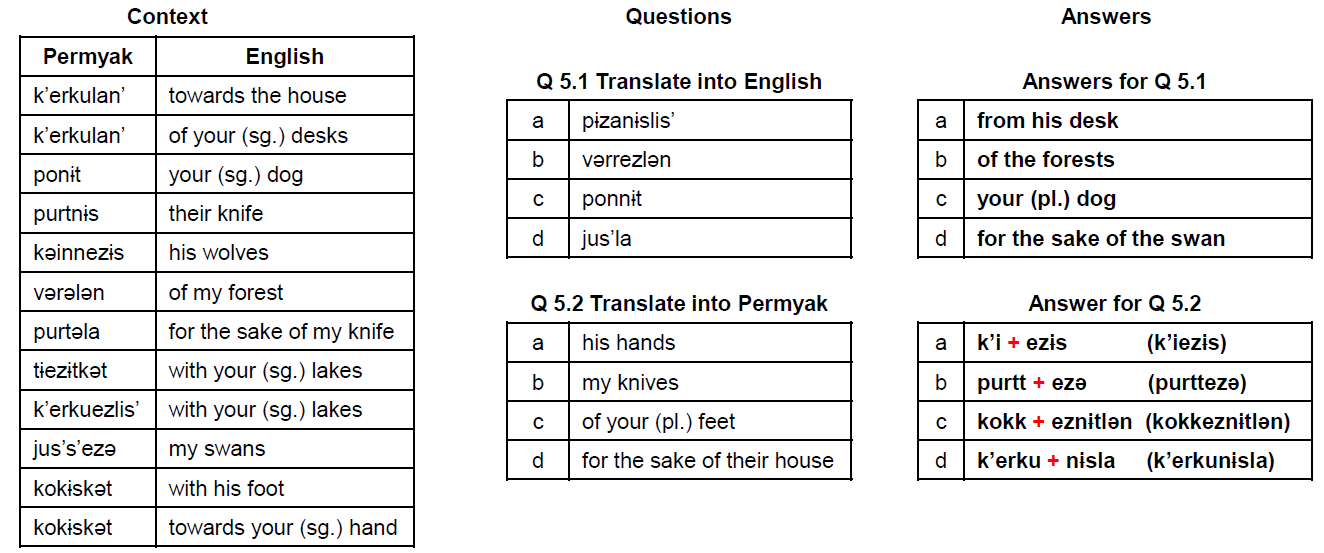}
 \caption{Partial context of the Permyak puzzle (full puzzle in App.~\ref{sec:appExamples}, Fig.~\ref{fig:Permyak}). Each Permyak word corresponds to a sequence of two to six English tokens, reflecting the language's agglutinative morphology.}
  \label{fig:Permyak_example}
\end{figure*}

This appendix provides an example to support the choice of whitespace tokenization.

\subsection{Why whitespace tokenization?}

A whitespace token in one language can correspond to several morphemes that recur independently. Fig.~\ref{fig:Permyak_example} shows the Permyak case: each Permyak word maps to two-to-six English words because Permyak is agglutinative, with suffixes for case, number, possession, mode, and time stacked on a stem. A sub-word segmenter would expose this structure and reveal that some apparently singleton Permyak tokens share morphemes with other context examples. 

We nonetheless adopt whitespace tokenization for two reasons. (1) it is language-agnostic: our corpus of 53 puzzles spans 52 languages from 34 families, and any sub-word segmentation would require language-specific resources, contradicting the self-contained nature of linguistic puzzles. (2) whitespace tokenization yields a \emph{conservative} estimate of the damage score. If a whitespace token is a singleton, every sub-token morpheme inside it is also a singleton. The converse is false: a sub-word morpheme can be singleton at the morpheme level even when its enclosing whitespace token is not. Whitespace damage therefore lower-bounds morpheme-level damage, and the structural unsolvability claim ($D_Q > 0 \Rightarrow$ unsolvable under whitespace-token reasoning) holds under any finer segmentation. Thus, we may classify some morpheme-recoverable puzzles as fragile; the LLM-as-judge experiment catches the cases where this happens.

This concern is not hypothetical:~\citet{cunveiling} show that making morpheme boundaries explicit measurably improves LLM solvability of Linguistics Olympiad puzzles, confirming that whitespace tokens and the units models actually reason over can come apart. Our choice of whitespace tokens is therefore deliberately conservative rather than assumed to be neutral.



\section{Prompts}
\label{sec:appPrompts}
The LLMs used in the experiments are GPT-5.4 (Version: \texttt{gpt-5.4-2026-03-05}), Gemini 3 Flash (Version: \texttt{gemini-3-flash-preview}), and Claude Sonnet 4.6 (Version: \texttt{claude-sonnet-4-6}). They were accessed on 28th March 2026.
\subsection{Puzzle Solving with Abstention}
\label{subsec:appPromptsExp1}

\begin{tcolorbox}[colback=blue!5!white, colframe=blue!75!black, title=System Prompt: Puzzle Solving with Abstention, fontupper=\sffamily]
You are an expert linguist and puzzle solver. The puzzle given below has a PREAMBLE, CONTEXT, and QUESTIONS. \\
Your task:\\
1. Review the provided PREAMBLE, CONTEXT, and QUESTIONS.\\
2. Solve each question strictly and entirely using the information provided in the PREAMBLE and CONTEXT sections. If the question cannot be solved, then you are strictly supposed to output "N/A".\\
3. For each question, provide your derived answer, and a "backtrace" consisting of the numerical indices from the CONTEXT examples you used to derive that answer.\\
\end{tcolorbox}

\begin{tcolorbox}[colback=gray!5!white, colframe=gray!75!black, title=User Prompt: Puzzle Solving with Abstention, fontupper=\sffamily]
Analyze the puzzle below.\\
**---PREAMBLE---**\\
\{preamble\}\\ \\
**---CONTEXT---**\\
\{context\}\\ \\
**---QUESTIONS---**\\
\{all\_subquestions\}
\end{tcolorbox}

\clearpage
\subsection{LLM-as-a-Judge Puzzle Validity Assessment}
\label{subsec:appPromptsExp2}
\begin{tcolorbox}[colback=blue!5!white, colframe=blue!75!black, title=System Prompt: LLM-as-a-Judge Puzzle Validity Assessment, fontupper=\sffamily]
You are an expert linguistics auditor.
\end{tcolorbox}

\begin{tcolorbox}[colback=gray!5!white, colframe=gray!75!black, title=User Prompt: LLM-as-a-Judge Puzzle Validity Assessment, fontupper=\sffamily]
The following puzzle is the original puzzle, with the solutions to its questions.\\
**---PREAMBLE---**\\
\{preamble\}\\ \\
**---CONTEXT---**\\
\{context\}\\ \\
**---QUESTIONS---**\\
\{all\_subquestions\}\\ \\
**---ANSWERS---**\\
\{all\_subquestions\_answers\}\\ \\
Here is the modified version of the puzzle:\\
**---PREAMBLE---**\\
\{preamble\}\\ \\
**---CONTEXT---**\\
\{modified\_context\}\\ \\
**---QUESTIONS---**\\
\{all\_subquestions\}\\ \\
Is the answer to the original puzzle valid for the modified one, or the modified puzzle is different enough to invalidate the solution.
\end{tcolorbox}


\section{No-context Score}
\label{sec:appNoContext}

Table~\ref{tab:no_context_scores} shows the performance of puzzle solving and the abstention across three models on puzzles without the context examples.

\begin{table}[h]
\centering
\small
\begin{tabular}{lcc}
\toprule
\textbf{Model} & \textbf{Avg. EM} & \textbf{\#N/A} \\
\midrule
GPT & 0.173 & 10 \\
Claude & 0.151 & 4 \\
Gemini & 0.270 & 0 \\
\bottomrule
\end{tabular}
\caption{\textbf{Avg. EM} are the average Exact Match (EM) scores across the three models in no-context versions; \textbf{\#N/A} counts puzzles where the model abstained or produced no output.}
\label{tab:no_context_scores}
\end{table}



\section{Distribution of UKLO Puzzle Formats}
\label{sec:appFormats}
There are several established formats of linguistic puzzles, including Rosetta Stone, Match-Up (also known as Chaos), Monolingual, Pattern, Computational, and Text. The definitions below are taken \emph{verbatim} from the UKLO website.\footnote{\url{https://www.uklo.org/technical-information/\#qformat}} These are the formats that are assigned to the puzzles by their authors.

\begin{itemize}
\item \textbf{Rosetta}: The data are sets of corresponding words or phrases between different languages/writing systems, with most of the correspondences given. Parts may be omitted from the data set, leaving gaps to be filled. You must be able to give new correspondences (typically translations) to solve the task.
\item \textbf{Match-Up}: The data are sets of corresponding words or phrases in multiple languages/writing systems, but with few of the correspondences given. If some words are not part of a set, it can still count as a match-up. You must be able to give new correspondences (typically translations) to solve the task.
\item \textbf{Monolingual}: The data are texts in an unknown language (or equivalent), with no direct translation (or transliteration for writing systems) given, with the possible exception of 1-2 words. You must be able to translate from the language to solve the task.
\item \textbf{Pattern}: The data are words or sets of forms of words/cognates, conforming to a pattern (possibly with some exceptions). You must be able to give other words conforming to this pattern or identify outliers to solve the task, but (unlike in a Rosetta) there is no translation component.
\item \textbf{Computational}: The problem data includes a description of a computational or other logical system. To solve the problem, you must be able to analyse and implement this system.
\item \textbf{Text}: The data are whole texts presented in different languages or scripts, but not subdivided further. To solve the problem, you must use context and other clues to deduce linguistic rules.
\end{itemize}

Table~\ref{tbl:UKLOformats} contains the distribution of the 235 UKLO puzzles used between 2010 and 2025 across the six formats listed above. Puzzles are assigned to more than one format. We use the format as specified by UKLO. For example, the combination ['Rosetta', 'Match-up'] and ['Match-up', 'Rosetta'] are treated as two different combinations.

\begin{table}[htbp]
\centering
\small
\begin{tabular}{@{} l c @{}}
\toprule
\textbf{Format} & \textbf{Count} \\
\midrule
{['Rosetta']} & 106 \\
{['Match-up']} & 66 \\
{['Rosetta', 'Match-up']} & 3 \\
{['Match-up', 'Rosetta']} & 1 \\
\midrule
\textbf{Sum} & \textbf{176 (75\%)} \\
\midrule
{['Computational']} & 6 \\
{['Match-up', 'Monolingual']} & 1 \\
{['Match-up', 'Text']} & 1 \\
{['Monolingual', 'Match-up']} & 1 \\
{['Monolingual', 'Rosetta']} & 1 \\
{['Monolingual']} & 10 \\
{['Pattern', 'Match-up']} & 2 \\
{['Pattern', 'Rosetta']} & 1 \\
{['Pattern']} & 30 \\
{['Text', 'Match-up']} & 3 \\
{['Text']} & 3 \\
\midrule
\textbf{Grand Total} & \textbf{235} \\
\bottomrule
\end{tabular}
\caption{Counts of how many puzzles of a particular format (or combination of formats) were used in UKLO from 2010 to 2025.}
\label{tbl:UKLOformats}
\end{table}



\section{LLM Judge Verdicts}
\label{sec:appResultsTables}


This section provides the detailed, puzzle-by-puzzle results of the LLM-as-a-judge solvability assessment. Because the Question Damage Score ($D_Q$) relies on a deterministic structural argument, we use the LLM judge as an independent, language-aware empirical cross-check to verify whether the modified contexts are truly insufficient.

Table~\ref{tab:combined_verdicts_performance} presents the raw judge verdicts (Valid or Unsolvable) across all 53 puzzles for both the Random and ECC-inspired deletion strategies. Puzzles where the model's Exact Match (EM) score strictly increased despite the deletion are highlighted in bold.

To further contextualize these judgments against actual model performance, we isolate two distinct behavioral patterns:

Table~\ref{tab:combined_contamination} details the cases where a model's Exact Match score equaled or strictly exceeded its original score on a puzzle that the judge confirmed to be unsolvable ($\Delta \ge 0$). This table provides the item-level evidence for pretraining recall; the model successfully produces answers even when the derivation path has been provably broken.

Conversely, Table~\ref{tab:combined_derailment} highlights the inverse failure mode. It lists the cases where a puzzle remains structurally solvable according to the judge, yet the removal of a redundant context example causes a strict drop in performance ($\Delta < 0$). This illustrates the brittleness of the models' deductive capabilities, showing that even harmless deletions can derail their reasoning.

\begin{table}[h]
\centering
\small
\begin{tabular}{lcc}
\toprule
\textbf{Model} & \textbf{Wilcoxon $p$-value} & \textbf{McNemar $p$-value} \\
& \textbf{(Puzzle-level EM)} & \textbf{(Question-level Binary)} \\
\midrule
GPT & 0.4208 & 0.3211 \\
Claude & 0.5552 & 0.4799 \\
Gemini & \textbf{0.0052} & \textbf{0.0022} \\
\bottomrule
\end{tabular}
\vspace{0.1in}
\caption{Statistical significance tests comparing model performance on the Original vs. ECC-deleted puzzles. Bold values indicate statistical significance at $\alpha = 0.05$. The models are GPT-5.4, Claude Sonnet 4.6, and Gemini 3 Flash.}
\label{tab:significance}
\end{table}

\begin{table*}[htbp]
\centering
\small
\centering
\begin{tabular}{@{} l l c c c l l c c @{}}
\toprule
& & \multicolumn{2}{c}{\textbf{Judge Verdict}} & & & & \multicolumn{2}{c}{\textbf{Judge Verdict}} \\
\cmidrule(lr){3-4} \cmidrule(lr){8-9}
\textbf{ID} & \textbf{Name} & \textbf{Random} & \textbf{ECC} & \phantom{abc} & \textbf{ID} & \textbf{Name} & \textbf{Random} & \textbf{ECC} \\
\midrule
5 & \href{https://www.uklo.org/wp-content/uploads/2022/09/2010.5-Turkish.pdf}{Turkish} & Unsolv. & Unsolv. & & 139 & \href{https://www.uklo.org/wp-content/uploads/2022/05/2019_R2_1.-Afrihili.pdf}{Afrihili} & Valid & Unsolv. \\
16 & \href{https://www.uklo.org/wp-content/uploads/2022/09/2011.4-Ulwa.pdf}{Ulwa} & Valid & Valid & & 155 & \href{https://www.uklo.org/wp-content/uploads/2022/05/2020_R2_2-Yoruba.pdf}{Yoruba} & Valid & Valid \\
24 & \href{https://www.uklo.org/wp-content/uploads/2022/09/2011r2.5-Tadaksahak.pdf}{Tadaksahak} & Valid & Valid & & 160 & \href{https://www.uklo.org/wp-content/uploads/2022/05/2021_2-Kabyle.pdf}{Kabyle} & Valid & Unsolv. \\
25 & \href{https://www.uklo.org/wp-content/uploads/2022/09/2012.1-Yolmo.pdf}{Yolmo} & Valid & Unsolv. & & 173 & \href{https://www.uklo.org/wp-content/uploads/2022/05/2021_R2_3-Ainu.pdf}{Ainu} & Valid & Valid \\
40 & \href{https://www.uklo.org/wp-content/uploads/2022/09/2013.1-Yodaspeak.pdf}{Yodaspeak} & Valid & Valid & & 175 & \href{https://www.uklo.org/wp-content/uploads/2022/05/2021_R2_5-Tawala.pdf}{Tawala} & Unsolv. & Unsolv. \\
42 & \href{https://www.uklo.org/wp-content/uploads/2022/09/2013.3-Pali.pdf}{Pali} & Unsolv. & Unsolv. & & 176 & \href{https://www.uklo.org/wp-content/uploads/2022/05/1_UKLO-2022-Swedish_The-Pink-Pig-is-Pink_-Complete-Script.pdf}{Swedish} & Valid & Unsolv. \\
45 & \href{https://www.uklo.org/wp-content/uploads/2022/09/2013.5-Bulgarian.pdf}{Bulgarian} & Unsolv. & Unsolv. & & 183 & \href{https://www.uklo.org/wp-content/uploads/2022/06/8_Adv_UKLO-2022-Zuni_Zuni-Tunes__Complete-Script.pdf}{Zuni} & Valid & Valid \\
52 & \href{https://www.uklo.org/wp-content/uploads/2022/09/2013r2.3-Beja.pdf}{Beja} & Unsolv. & Unsolv. & & 184 & \href{https://www.uklo.org/wp-content/uploads/2022/05/9_Adv_UKLO-2022-Tseltal__Complete-Script.pdf}{Tseltal} & Valid & Valid \\
55 & \href{https://www.uklo.org/wp-content/uploads/2022/08/2014.1-Estonian.pdf}{Estonian} & \textbf{Unsolv.} & \textbf{Unsolv.} & & 188 & \href{https://www.uklo.org/wp-content/uploads/2022/05/2022_R2_3_Niuean.pdf}{Niuean} & \textbf{Valid} & Unsolv. \\
61 & \href{https://www.uklo.org/wp-content/uploads/2022/08/2014.7-Ilokano.pdf}{Ilokano} & Unsolv. & Valid & & 190 & \href{https://www.uklo.org/wp-content/uploads/2022/05/2022_R2_5_Taos.pdf}{Taos} & Valid & Unsolv. \\
66 & \href{https://www.uklo.org/wp-content/uploads/2022/08/2014r2.3-Yidiny.pdf}{Yidiny} & \textbf{Valid} & Unsolv. & & 193 & \href{https://www.uklo.org/wp-content/uploads/2023/03/2023_R1_3-Gilbertese.pdf}{Gilbertese} & Valid & Valid \\
67 & \href{https://www.uklo.org/wp-content/uploads/2022/08/2014r2.4-Navajo.pdf}{Navajo} & Valid & Valid & & 195 & \href{https://www.uklo.org/wp-content/uploads/2023/03/2023_R1_5-Permyak.pdf}{Permyak} & Valid & Valid \\
69 & \href{https://www.uklo.org/wp-content/uploads/2022/05/2015_1-Karelian.pdf}{Karelian} & \textbf{Unsolv.} & \textbf{Unsolv.} & & 197 & \href{https://www.uklo.org/wp-content/uploads/2023/03/2023_R1_7-Lardil.zip}{Lardil} & \textbf{Valid} & \textbf{Valid} \\
72 & \href{https://www.uklo.org/wp-content/uploads/2022/05/2015_4.-Old-English.pdf}{Old English} & Unsolv. & Unsolv. & & 199 & \href{https://www.uklo.org/wp-content/uploads/2023/03/2023_R1_9-Kiche.pdf}{Kiche} & Valid & Unsolv. \\
75 & \href{https://www.uklo.org/wp-content/uploads/2022/05/2015_7-Murrinhpatha.pdf}{Murrinhpatha} & \textbf{Valid} & \textbf{Valid} & & 201 & \href{https://www.uklo.org/wp-content/uploads/2023/03/2023_R2_1-Abawiri.pdf}{Abawiri} & \textbf{Valid} & \textbf{Valid} \\
88 & \href{https://www.uklo.org/wp-content/uploads/2022/05/2016_5-Amele.pdf}{Amele} & Valid & Unsolv. & & 203 & \href{https://www.uklo.org/wp-content/uploads/2023/04/2023_R2_3-Pular.pdf}{Pular} & Unsolv. & \textbf{Unsolv.} \\
92 & \href{https://www.uklo.org/wp-content/uploads/2022/05/2016_9.-Nhanda.pdf}{Nhanda} & Unsolv. & Valid & & 204 & \href{https://www.uklo.org/wp-content/uploads/2023/03/2023_R2_4-Komnzo.pdf}{Komnzo} & Valid & Valid \\
93 & \href{https://www.uklo.org/wp-content/uploads/2022/05/2016_10.-Nung.pdf}{Nung} & Unsolv. & Valid & & 205 & \href{https://www.uklo.org/wp-content/uploads/2023/03/2023_R2_5-Mongo.pdf}{Mongo} & Unsolv. & Unsolv. \\
94 & \href{https://www.uklo.org/wp-content/uploads/2022/05/2016_R2.1-Malay.pdf}{Malay} & Unsolv. & Unsolv. & & 209 & \href{https://www.uklo.org/wp-content/uploads/2024/04/2024_R1_4-Xhosa.pdf}{Xhosa} & Valid & Valid \\
103 & \href{https://www.uklo.org/wp-content/uploads/2022/05/2017_5.-Basque.pdf}{Basque} & Valid & Unsolv. & & 214 & \href{https://www.uklo.org/wp-content/uploads/2024/04/2024_R1_9-Zou.pdf}{Zou} & Valid & Unsolv. \\
106 & \href{https://www.uklo.org/wp-content/uploads/2022/05/2017_8.-Choctaw.pdf}{Choctaw} & \textbf{Valid} & \textbf{Unsolv.} & & 216 & \href{https://www.uklo.org/wp-content/uploads/2024/03/2024_R2_1-Yawalapiti.pdf}{Yawalapiti} & Valid & \textbf{Unsolv.} \\
109 & \href{https://www.uklo.org/wp-content/uploads/2022/05/2017_R2.1-Nepali.pdf}{Nepali} & Valid & Unsolv. & & 217 & \href{https://www.uklo.org/wp-content/uploads/2024/03/2024_R2_2-Taa.pdf}{Taa} & Unsolv. & Unsolv. \\
115 & \href{https://www.uklo.org/wp-content/uploads/2022/05/2018_2-Lithuanian.pdf}{Lithuanian} & Valid & Valid & & 219 & \href{https://www.uklo.org/wp-content/uploads/2024/03/2024-R2_4-Coptic.pdf}{Coptic} & Valid & Unsolv. \\
118 & \href{https://www.uklo.org/wp-content/uploads/2022/05/2018_5-Gilbertese.pdf}{Gilbertese} & Unsolv. & Unsolv. & & 223 & \href{https://www.uklo.org/wp-content/uploads/2025/04/2025R1-3-Saisiyat.pdf}{Saisiyat} & Unsolv. & Unsolv. \\
128 & \href{https://www.uklo.org/wp-content/uploads/2022/05/2018_R2_5-Mayangna.pdf}{Mayangna} & Unsolv. & Unsolv. & & 229 & \href{https://www.uklo.org/wp-content/uploads/2025/04/2025R1-9-Cherokee.pdf}{Cherokee} & \textbf{Valid} & \textbf{Unsolv.} \\
132 & \href{https://www.uklo.org/wp-content/uploads/2022/05/2019_4-Welsh.pdf}{Welsh} & Unsolv. & Unsolv. & & 235 & \href{https://www.uklo.org/wp-content/uploads/2025/04/2025_R2_5-Kavalan.pdf}{Kavalan} & Valid & \textbf{Unsolv.} \\
137 & \href{https://www.uklo.org/wp-content/uploads/2022/05/2019_9-Ndebele.pdf}{Ndebele} & Unsolv. & Unsolv. & & & & & \\
\bottomrule
\end{tabular}
\caption{Breakdown of LLM judge verdicts across the Random and ECC-inspired strategies. \textbf{Bold} indicates cases where the Exact Match score on the modified puzzle is strictly higher than the score on the original puzzle.}
\label{tab:combined_verdicts_performance}
\end{table*}

\begin{table*}[htbp]
\centering
\small
\centering
\begin{tabular}{@{} l l c c c c c @{}}
\toprule
& & & \multicolumn{2}{c}{\textbf{ECC-inspired}} & \multicolumn{2}{c}{\textbf{Random}} \\
\cmidrule(lr){4-5} \cmidrule(lr){6-7}
\textbf{ID} & \textbf{Name} & \textbf{Original Score} & \textbf{Score} & \textbf{$\Delta_{\text{ecc} - \text{og}}$} & \textbf{Score} & \textbf{$\Delta_{\text{rand} - \text{og}}$} \\
\midrule
5   & \href{https://www.uklo.org/wp-content/uploads/2022/09/2010.5-Turkish.pdf}{Turkish} & 0.860 & 0.860 & +0.000 & 0.860 & +0.000 \\
42  & \href{https://www.uklo.org/wp-content/uploads/2022/09/2013.3-Pali.pdf}{Pali} & 0.880 & 0.880 & +0.000 & 0.880 & +0.000 \\
52  & \href{https://www.uklo.org/wp-content/uploads/2022/09/2013r2.3-Beja.pdf}{Beja} & 0.360 & 0.360 & +0.000 & -- & -- \\
55  & \href{https://www.uklo.org/wp-content/uploads/2022/08/2014.1-Estonian.pdf}{Estonian} & 0.900 & 1.000 & +0.100 & 1.000 & +0.100 \\
61  & \href{https://www.uklo.org/wp-content/uploads/2022/08/2014.7-Ilokano.pdf}{Ilokano} & 1.000 & -- & -- & 1.000 & +0.000 \\
66  & \href{https://www.uklo.org/wp-content/uploads/2022/08/2014r2.3-Yidiny.pdf}{Yidiny} & 0.920 & 0.920 & +0.000 & -- & -- \\
69  & \href{https://www.uklo.org/wp-content/uploads/2022/05/2015_1-Karelian.pdf}{Karelian} & 0.310 & 0.380 & +0.070 & 0.540 & +0.230 \\
72  & \href{https://www.uklo.org/wp-content/uploads/2022/05/2015_4.-Old-English.pdf}{Old English} & 1.000 & -- & -- & 1.000 & +0.000 \\
92  & \href{https://www.uklo.org/wp-content/uploads/2022/05/2016_9.-Nhanda.pdf}{Nhanda} & 0.920 & -- & -- & 0.920 & +0.000 \\
103 & \href{https://www.uklo.org/wp-content/uploads/2022/05/2017_5.-Basque.pdf}{Basque} & 0.860 & 0.860 & +0.000 & -- & -- \\
106 & \href{https://www.uklo.org/wp-content/uploads/2022/05/2017_8.-Choctaw.pdf}{Choctaw} & 0.800 & 0.900 & +0.100 & -- & -- \\
128 & \href{https://www.uklo.org/wp-content/uploads/2022/05/2018_R2_5-Mayangna.pdf}{Mayangna} & 0.170 & 0.170 & +0.000 & -- & -- \\
132 & \href{https://www.uklo.org/wp-content/uploads/2022/05/2019_4-Welsh.pdf}{Welsh} & 1.000 & 1.000 & +0.000 & 1.000 & +0.000 \\
176 & \href{https://www.uklo.org/wp-content/uploads/2022/05/1_UKLO-2022-Swedish_The-Pink-Pig-is-Pink_-Complete-Script.pdf}{Swedish} & 1.000 & 1.000 & +0.000 & -- & -- \\
203 & \href{https://www.uklo.org/wp-content/uploads/2023/04/2023_R2_3-Pular.pdf}{Pular} & 0.560 & 0.670 & +0.110 & -- & -- \\
216 & \href{https://www.uklo.org/wp-content/uploads/2024/03/2024_R2_1-Yawalapiti.pdf}{Yawalapiti} & 0.550 & 0.590 & +0.040 & -- & -- \\
217 & \href{https://www.uklo.org/wp-content/uploads/2024/03/2024_R2_2-Taa.pdf}{Taa} & 0.290 & 0.290 & +0.000 & -- & -- \\
219 & \href{https://www.uklo.org/wp-content/uploads/2024/03/2024-R2_4-Coptic.pdf}{Coptic} & 0.380 & 0.380 & +0.000 & -- & -- \\
223 & \href{https://www.uklo.org/wp-content/uploads/2025/04/2025R1-3-Saisiyat.pdf}{Saisiyat} & 0.000 & 0.000 & +0.000 & 0.000 & +0.000 \\
229 & \href{https://www.uklo.org/wp-content/uploads/2025/04/2025R1-9-Cherokee.pdf}{Cherokee} & 0.700 & 0.900 & +0.200 & -- & -- \\
235 & \href{https://www.uklo.org/wp-content/uploads/2025/04/2025_R2_5-Kavalan.pdf}{Kavalan} & 0.440 & 0.560 & +0.120 & -- & -- \\
\bottomrule
\end{tabular}
\caption{Puzzles judged unsolvable by Gemini where Exact Match performance increased or remained equal to the original ($\Delta \geq 0$). Dashes (--) indicate the puzzle was either not judged unsolvable under that strategy or its score dropped.}
\label{tab:combined_contamination}
\end{table*}

\begin{table*}[htbp]
\centering
\small
\centering
\begin{tabular}{@{} l l c c c c c @{}}
\toprule
& & & \multicolumn{2}{c}{\textbf{ECC-inspired}} & \multicolumn{2}{c}{\textbf{Random}} \\
\cmidrule(lr){4-5} \cmidrule(lr){6-7}
\textbf{ID} & \textbf{Name} & \textbf{Original Score} & \textbf{Score} & \textbf{$\Delta_{\text{ecc} - \text{og}}$} & \textbf{Score} & \textbf{$\Delta_{\text{rand} - \text{og}}$} \\
\midrule
16  & \href{https://www.uklo.org/wp-content/uploads/2022/09/2011.4-Ulwa.pdf}{Ulwa} &  1.000 & 0.600 & -0.400 & -- & -- \\
24  & \href{https://www.uklo.org/wp-content/uploads/2022/09/2011r2.5-Tadaksahak.pdf}{Tadaksahak} & 0.890 & 0.670 & -0.220 & 0.780 & -0.110 \\
40  & \href{https://www.uklo.org/wp-content/uploads/2022/09/2013.1-Yodaspeak.pdf}{Yodaspeak} & 1.000 & 0.900 & -0.100 & -- & -- \\
93  & \href{https://www.uklo.org/wp-content/uploads/2022/05/2016_10.-Nung.pdf}{Nung} & 0.120 & 0.000 & -0.120 & -- & -- \\
109 & \href{https://www.uklo.org/wp-content/uploads/2022/05/2017_R2.1-Nepali.pdf}{Nepali} & 0.900 & -- & -- & 0.700 & -0.200 \\
115 & \href{https://www.uklo.org/wp-content/uploads/2022/05/2018_2-Lithuanian.pdf}{Lithuanian} & 1.000 & 0.880 & -0.120 & 0.880 & -0.120 \\
176 & \href{https://www.uklo.org/wp-content/uploads/2022/05/1_UKLO-2022-Swedish_The-Pink-Pig-is-Pink_-Complete-Script.pdf}{Swedish} & 1.000 & -- & -- & 0.830 & -0.170 \\
183 & \href{https://www.uklo.org/wp-content/uploads/2022/06/8_Adv_UKLO-2022-Zuni_Zuni-Tunes__Complete-Script.pdf}{Zuni} & 0.900 & -- & -- & 0.800 & -0.100 \\
184 & \href{https://www.uklo.org/wp-content/uploads/2022/05/9_Adv_UKLO-2022-Tseltal__Complete-Script.pdf}{Tseltal} & 0.750 & 0.670 & -0.080 & -- & -- \\
190 & \href{https://www.uklo.org/wp-content/uploads/2022/05/2022_R2_5_Taos.pdf}{Taos} & 0.380 & -- & -- & 0.250 & -0.130 \\
199 & \href{https://www.uklo.org/wp-content/uploads/2023/03/2023_R1_9-Kiche.pdf}{Kiche} & 0.670 & -- & -- & 0.560 & -0.110 \\
204 & \href{https://www.uklo.org/wp-content/uploads/2023/03/2023_R2_4-Komnzo.pdf}{Komnzo} & 0.270 & -- & -- & 0.180 & -0.090 \\
214 & \href{https://www.uklo.org/wp-content/uploads/2024/04/2024_R1_9-Zou.pdf}{Zou} & 0.730 & -- & -- & 0.640 & -0.090 \\
216 & \href{https://www.uklo.org/wp-content/uploads/2024/03/2024_R2_1-Yawalapiti.pdf}{Yawalapiti} & 0.550 & -- & -- & 0.320 & -0.230 \\
\bottomrule
\end{tabular}
\caption{Puzzles judged solvable by Gemini where Exact Match performance strictly dropped ($\Delta < 0$). Dashes (--) indicate the puzzle was either not judged solvable under that strategy or its score did not drop.}
\label{tab:combined_derailment}
\end{table*}


\clearpage
\section{Statistical Significance Tests}
\label{sec:appStatSign}
To evaluate the impact of context deletion on model performance, we utilized two distinct statistical tests. Because we evaluated the exact same set of $N=53$ puzzles under both the original and modified conditions, our data is inherently paired (dependent).

\paragraph{Wilcoxon Signed-Rank Test}
The Wilcoxon signed-rank test is a non-parametric statistical hypothesis test used to compare two related samples. We select this test for the puzzle-level Exact Match (EM) scores. Because the scores are continuous values and the sample size is small ($N=53$), we cannot reliably assume that the differences in scores are normally distributed. The Wilcoxon test does not require normal distribution, making it the mathematically appropriate alternative to a standard paired t-test for measuring whether the magnitude of performance drops is significant.

\paragraph{McNemar's Test}
McNemar's test is a non-parametric test designed specifically for paired binary data. We select this test for the question-level analysis because the model outcomes at this resolution are strictly binary (1 for a correct answer, 0 for an incorrect answer). Instead of looking at averages, McNemar's test isolates the pairs where the specific questions a model got right in one condition but wrong in the other. This determines if the shift in accuracy is a statistically significant effect of the deleted context rather than random variance.
%



\end{document}